\documentclass[runningheads]{llncs}

\usepackage{eccv}

\usepackage{eccvabbrv}

\usepackage{graphicx}
\usepackage{booktabs}
\usepackage[table]{xcolor}
\usepackage{multirow}
\usepackage{amsmath}

\usepackage[accsupp]{axessibility}

\usepackage{hyperref}

\usepackage{orcidlink}

\begin{document}

\title{High-Resolution Artwork Outpainting with Global Blueprint Guidance and Layout Control} 

\titlerunning{Blueprint-Guided Artwork Outpainting}

\author{Junha Kim\orcidlink{0009-0003-0939-0113} \and
Hyunjoon Park\orcidlink{0009-0008-1860-3172} \and
Donghyeon Cho\textsuperscript{\ensuremath{\dagger}}\orcidlink{0000-0002-2184-921X}}

\authorrunning{J.~Kim et al.}

\institute{Hanyang University, Seoul, Republic of Korea\\
\email{\{poohoh,junippini83,doncho\}@hanyang.ac.kr}}

\maketitle

\begingroup
\renewcommand{\thefootnote}{\ensuremath{\dagger}}
\footnotetext{Corresponding author.}
\endgroup

\begin{abstract}
Image outpainting extends an image beyond its original borders, requiring seamless style integration and globally coherent scene completion.
Building on the success of diffusion models, recent methods have achieved substantial improvements in visual quality.
In practice, however, high-resolution outpainting is commonly performed via progressive expansion around a fixed source image, particularly in artwork scenarios.
Despite this progress, existing approaches still suffer from three key limitations: (i) the absence of a reliable global planning mechanism, which leads to structural instability and error accumulation at high resolutions; (ii) limited spatial controllability beyond text prompts, making it difficult to place objects at user-specified locations; and (iii) high inference latency caused by inherently sequential patch generation.
To address these issues, we propose a global blueprint-guided two-stage diffusion framework for layout-controllable high-resolution outpainting with efficient parallel synthesis.
In Stage~1, we generate a low-resolution global blueprint using a layout adapter that injects bounding-box conditions into a Stable Diffusion inpainting backbone, producing a globally consistent structural plan while extracting global guidance features.
In Stage~2, we synthesize high-resolution local patches in parallel by injecting the blueprint-derived global guidance and initializing each patch from the blueprint using the low-frequency preservation property of forward diffusion.
This design eliminates sequential dependency while maintaining global coherence.
Extensive experiments on large-scale artwork datasets demonstrate improved visual fidelity, stronger semantic consistency, and substantially reduced inference time compared to prior baselines, while uniquely supporting explicit layout control for artwork outpainting.
Our code will be available at \url{https://github.com/poohoh/BlueOut}.
\end{abstract}

\section{Introduction}
\label{sec:intro}

Image outpainting extends an image beyond its original borders.
The key challenge lies in seamless integration, which requires matching the style, color, and context of the source image rather than merely enlarging the canvas.
Although outpainting is often considered analogous to inpainting, it is inherently more challenging.
In inpainting, the target region is surrounded by known pixels, which facilitates reliable context inference.
In contrast, outpainting synthesizes content in unbounded regions outside the image, introducing substantially greater uncertainty.
Consequently, successful outpainting requires not only plausible local synthesis but also an accurate understanding of the global context to maintain structural naturalness and contextual consistency.
Moreover, in artwork outpainting for immersive display applications such as large-scale projection and VR/AR environments~\cite{vangogh_expo}, legacy paintings must be adapted to new canvas sizes while preserving their original visual fidelity and aspect ratios.
This requirement makes the process highly sensitive to stylistic distortion and compositional imbalance, thereby making accurate global context modeling particularly critical.

Building on the success of diffusion models, recent outpainting methods have achieved substantial improvements in visual quality.
In practice, most existing approaches operate at fixed resolutions, typically $512 \times 512$ or $256 \times 256$, which correspond to the standard output sizes of diffusion backbones~\cite{alignnoise,lightsout,powerpaint,prefpaint}.
To enable high-resolution expansion, several progressive strategies have been proposed that iteratively extend the image by sliding a local window across the canvas~\cite{proout,nuwa_infinity,very_long_natural}.
Although diffusion models have been successfully applied to progressive outpainting, existing methods still suffer from several limitations.

\begin{figure*}[t]
    \centering
    \includegraphics[width=1\linewidth]{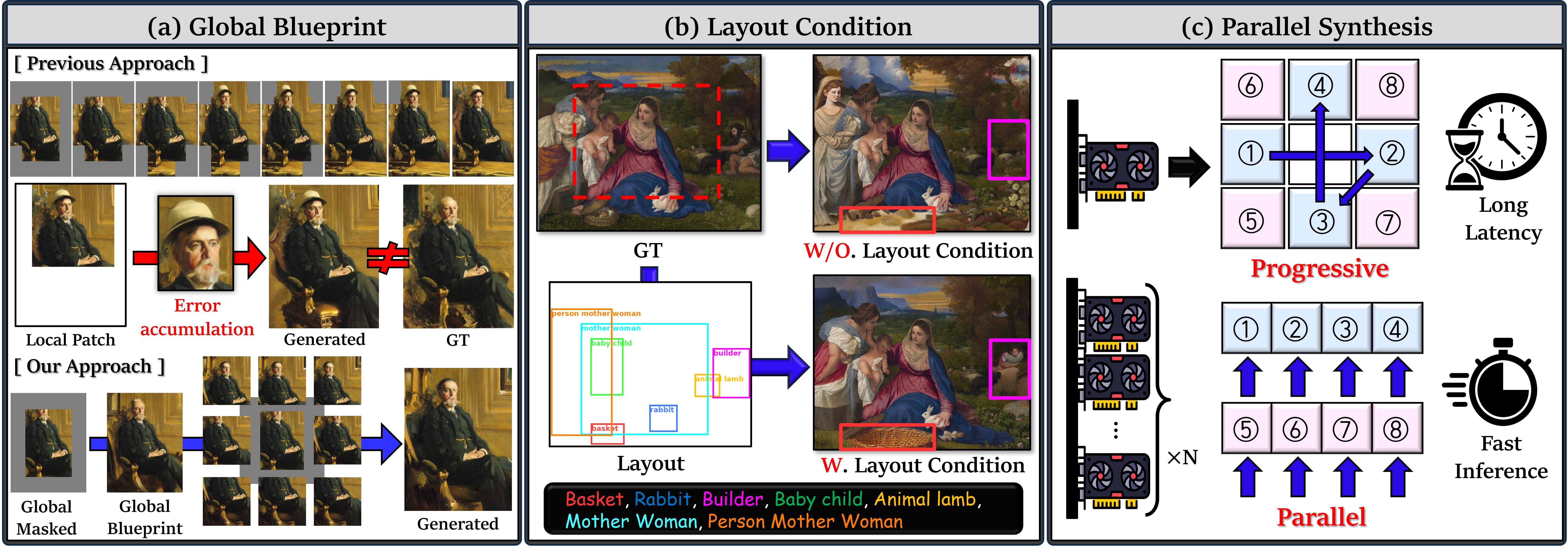}
    \caption{\textbf{Challenges in High-resolution Artwork Outpainting.}
    \textbf{(a)} Progressive methods accumulate errors without global planning; our blueprint-guided approach maintains global coherence.
    \textbf{(b)} Text-only conditioning struggles with object placement; our layout-conditioned framework enables spatial control.
    \textbf{(c)} Sequential generation incurs high latency; our parallel synthesis achieves up to $2.4 \times$ speedup.
    }
    \label{fig:intro}
\end{figure*}

As illustrated in Figure~\ref{fig:intro}(a), the first issue is the absence of a dedicated planning stage that provides reliable global blueprint for progressive high-resolution outpainting.
For example, ProOut~\cite{proout} injects a global hint image as a condition; however, it reflects only the intermediate canvas during progressive expansion rather than a predetermined global composition.
As a result, the method still relies on sequential generation and stitching of local windows.
As noted by prior studies~\cite{horizon,m3ddm}, this design leads to error accumulation, where inconsistencies from earlier stages propagate to later ones.
Consequently, without an explicit global blueprint, sequential expansion becomes sensitive to generation order and risks degrading global coherence in high-resolution results.
To fundamentally mitigate sequential error accumulation, we introduce a global blueprint-guided two-stage framework.
We first establish a global structural guideline by generating a low-resolution blueprint.
Based on this, we synthesize high-resolution details, preventing error propagation and ensuring stable quality even at scale.

As shown in Figure~\ref{fig:intro}(b), the second issue is the lack of a mechanism in current state-of-the-art outpainting models to precisely control generated content through layout conditions.
Before diffusion models become mainstream, early generative approaches, such as those utilizing GANs, attempt to generate specific objects or scenes at desired locations.
They achieve this by using layouts~\cite{scene_graph_expansion,boosting_image_outpainting,controllable_and_progressive} or sketches~\cite{sketch_guided_scenary,rego} as conditions, or by disentangling the structural semantics of an image during training~\cite{dhg_gan}.
However, these approaches rely on outdated backbones, yielding generation quality far below state-of-the-art techniques.
Among more recent diffusion-based studies, there have been attempts to control generation using text prompts~\cite{vip}.
While text-based conditions can specify the overall style or object presence, they cannot precisely control fine-grained layouts, such as exact positions or sizes.
Thus, text alone cannot ensure intended spatial arrangements.
In this paper, we model a mechanism that injects layout conditions, enabling accurate generation at user-specified locations while preserving the powerful generation capability of diffusion models.

The final issue, highlighted in Figure~\ref{fig:intro}(c), is the extreme computational inefficiency and long inference latency caused by sequential generation.
Diffusion models require iterative denoising, and progressive methods split the entire canvas and repeat sequential generation for high-resolution expansion, thus they inherently incur high computational costs even when generating a single image and runtime grows with the number of patches.
Moreover, this sequential dependency limits hardware utilization.
Due to the structural constraint where prior window generation must be completed before the next can begin, parallel processing is impossible even with sufficient memory.
This inevitably acts as a bottleneck for the entire system.
Recently, in video outpainting, Follow-Your-Canvas~\cite{follow_your_canvas} attempts to accelerate high-resolution generation by dividing the canvas into multiple spatial windows and decoding in parallel.
This method adopts a strategy similar to MultiDiffusion~\cite{multidiffusion}, blending overlaps to mitigate boundary artifacts.
However, overlap blending only smooths pixel transitions and fails to control global semantics.
This motivates our blueprint-guided coordination design, which uses the global blueprint as a shared structural guide for parallel high-resolution synthesis rather than relying on overlap blending alone.

Our main contributions can be summarized as follows:
\begin{itemize}
    \item \textbf{Blueprint-Guided Coordination Design:} We use the low-resolution global blueprint as a shared structural guide for high-resolution local synthesis, reducing error accumulation from sequential generation.
    \item \textbf{Layout-Controllable Framework:} We incorporate explicit layout conditioning in the planning stage, enabling users to specify object locations and descriptions in the expanded regions.
    \item \textbf{Efficient Parallel Synthesis:} We enable parallel patch generation for high-resolution artwork outpainting, reducing the sequential dependency of progressive methods and supporting multi-GPU inference.
\end{itemize}

\section{Related Works}
This section reviews prior work most relevant to our study, including outpainting, layout-controllable generation, and reference-guided synthesis.

\noindent \textbf{Outpainting.}
Early outpainting studies~\cite{painting_outside_the_box,boundless,inout,very_long_natural,generalized_image_outpainting,outpainting_by_queries,edge_guided_progressively,semie} primarily relied on GANs. 
While several works utilized layouts, sketches, or scene graphs to control object placement~\cite{sketch_guided_scenary,scene_graph_expansion,controllable_and_progressive}, they inherently suffered from limited generation quality and diversity due to their outdated backbones.
Recently, diffusion models have become the mainstream, driven by their superior synthesis quality. 
For instance, Stable Diffusion~(SD) Inpainting~\cite{stable_diffusion} enabled inpainting and outpainting via fine-tuning with random masks, and PowerPaint~\cite{powerpaint} introduced learnable task prompts for specific control. 
Other methods explored human aesthetic alignment~\cite{prefpaint}, context-aware outpainting via multimodal large language models~\cite{vip}, and training-free noise optimization~\cite{alignnoise}.
However, most diffusion-based methods operate at fixed resolutions (e.g., $512 \times 512$) and rely solely on text prompts, lacking fine-grained layout control. 
While some studies achieved continuous resolution expansion~\cite{pqdiff,easyoutpainter}, they cannot incorporate multimodal conditions. 
Although ProOut~\cite{proout} proposed progressive high-resolution expansion, it still suffers from severe inference latency and error accumulation due to sequential window generation.

\smallskip \noindent \textbf{Layout-Controllable Generation.}
To overcome the limited spatial controllability of standard text-to-image models, explicit layout conditioning has been extensively studied. 
Methods such as GLIGEN~\cite{gligen}, ControlNet~\cite{controlnet}, LayoutDiffusion~\cite{layoutdiffusion}, InstanceDiffusion~\cite{instancediffusion}, and MIGC~\cite{migc,migc++} inject structural conditions, such as bounding boxes, masks, or poses, to guide object synthesis at user-specified locations.
However, these approaches are primarily restricted to general image generation or local editing tasks. 
To the best of our knowledge, applying such advanced layout-control mechanisms to artwork outpainting, particularly to govern the global composition of extended regions, has not been thoroughly explored.

\smallskip \noindent \textbf{Reference-Guided Synthesis.}
In both image and video generation, reference-guided synthesis methods~\cite{ip_adapter,magic_animate,champ,idm_vton,ootdiffusion} aim to preserve the appearance or style of a reference image during generation.
This is typically achieved by extracting and injecting features via a supplementary encoder, such as ReferenceNet~\cite{animate_anyone}. 
While these approaches predominantly focus on transferring the visual appearance of specific objects, we fundamentally reinterpret this mechanism. 
In our framework, we leverage it to propagate the overall structural blueprint from the planning stage into the local patch generation process, thereby ensuring global structural and contextual consistency in high-resolution artwork outpainting.

\section{Method}
\subsection{Preliminaries}
\noindent \textbf{Latent Diffusion Models.}
Latent diffusion models (LDMs)~\cite{stable_diffusion} perform diffusion in a compressed latent space to improve computational efficiency.
Given an input image $x$, a pre-trained encoder $\mathcal{E}$ maps it to a latent representation $z_0 = \mathcal{E}(x)$.
In the forward process, Gaussian noise is gradually added according to a variance schedule $\{\beta_t\}_{t=1}^{T}$, yielding:
\begin{equation}
    z_t = \sqrt{\bar{\alpha}_t}\,z_0 + \sqrt{1-\bar{\alpha}_t}\,\epsilon, \quad \epsilon \sim \mathcal{N}(0,I),
\label{eq:forward}
\end{equation}
where $\alpha_t = 1-\beta_t$ and $\bar{\alpha}_t = \prod_{s=1}^{t}\alpha_s$.
At inference, denoising starts from \mbox{$z_T\sim\mathcal{N}(0,I)$} and iteratively recovers $z_0$, which is decoded via decoder $\mathcal{D}$.

\medskip
\noindent \textbf{Attention-Guided Noise Optimization.}
Prior to denoising in Stage~1, we optimize the initial noise by analyzing the self-attention maps of U-Net during the first denoising step.
Specifically, we adjust the noise so that spatial tokens in the target region sufficiently attend to those in the source region, following AlignNoise~\cite{alignnoise}.
This mitigates the common failure where the generated region becomes semantically disconnected from the original image.

\medskip
\noindent \textbf{Low-Frequency Preservation in Forward Diffusion.}
In standard diffusion theory, the final timestep $T$ is often regarded as pure Gaussian noise.
However, recent analyses~\cite{common_diffusion_noise,freeinit} show that this assumption does not strictly hold under typical noise schedules.
In particular, FreeInit demonstrates that while high-frequency components (e.g., textures and edges) are rapidly attenuated during early diffusion steps, low-frequency components (e.g., global layout and object placement) decay much more slowly and can remain partially preserved even at timestep $T$.
This low-frequency preservation implies that the macroscopic structure of an image can survive heavy noise corruption.
We exploit this property in our framework.
Specifically, we apply forward diffusion to the global blueprint generated in Stage~1, obtaining a noisy latent in which fine details are suppressed while global structural information is retained.
For each local patch in Stage~2, we crop the corresponding region from this noisy latent and use it as the initial noise.
As a result, all patches begin denoising from structurally aligned noise, enabling parallel synthesis while preserving global coherence.

\subsection{Overall Architecture}
\begin{figure}[!t]
    \centering
    \includegraphics[width=1\linewidth]{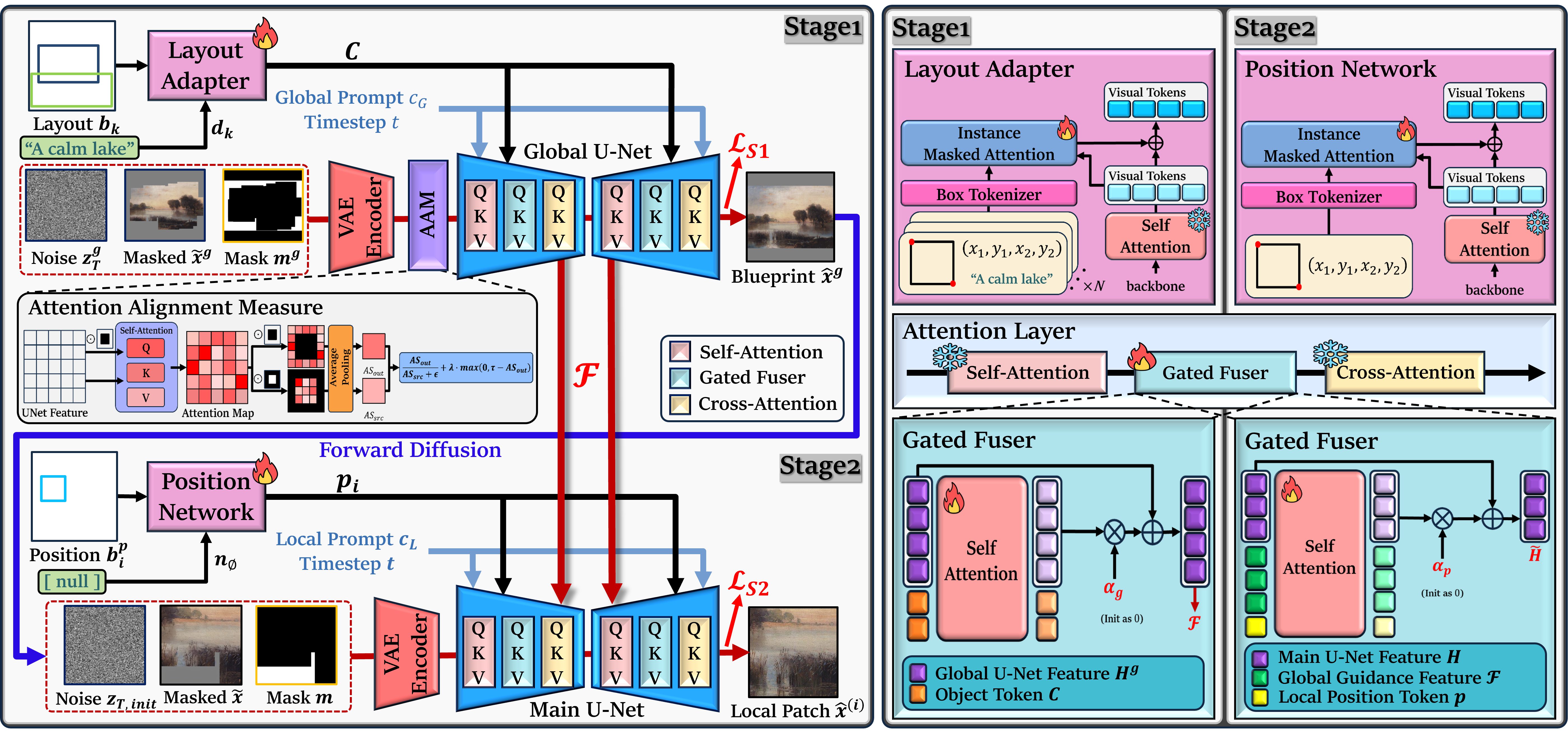}
    \caption{\textbf{Overview of Our Two-stage Framework.} 
    In Stage~1, the Attention Alignment Measure (AAM) first optimizes the initial noise to prevent semantic disconnection from the source image.
    The Layout Adapter then encodes user-specified bounding boxes $b_k$ and object descriptions $d_k$ into instance tokens $\mathrm{C}$.
    The Gated Fuser injects $\mathrm{C}$ into the global U-Net to generate a low-resolution blueprint $\hat{x}^g$ while exporting guidance features $\mathcal{F} = \{F_t^g\}_{t=1}^{T}$.
    In Stage~2, each high-resolution patch is initialized from $\hat{x}^g$ via forward diffusion and denoised in parallel.
    The Position Network encodes the patch's bounding box $b^p_i$ and a learnable null token $n_{\varnothing}$ into a position token $p_i$, and the Gated Fuser injects $\mathcal{F}$ and $p_i$ into the main U-Net.
    Module details are shown on the right.
    }
    \label{fig:pipeline}
\end{figure}

Our framework follows a global blueprint-guided two-stage pipeline, as illustrated in Figure~\ref{fig:pipeline}.
The core idea is to decouple global structural planning from high-resolution local synthesis.
The global blueprint then serves as a shared structural guide for parallel local synthesis, reducing error accumulation from sequential generation.
Stage~1 performs outpainting at low resolution over the full canvas to produce a global blueprint, simultaneously exporting timestep-wise guidance features.
Stage~2 then synthesizes high-resolution local patches in parallel, each initialized from the blueprint and guided by the exported features.
This design eliminates the sequential dependency of progressive methods while preserving global coherence.

We denote the Stage~1 blueprint generator as $G_{\theta_1}$ and the Stage~2 local patch synthesizer as $P_{\theta_2}$.
The overall pipeline is then formulated as follows.
In Stage~1, $G_{\theta_1}$ produces a global blueprint and guidance features:
\begin{equation}
    (\hat{x}^{g},\; \mathcal{F}) = G_{\theta_1}(z_T^{g},\, \tilde{x}^{g},\, m^{g},\, \mathcal{B},\, c_G),
\end{equation}
where $\hat{x}^{g}$ is the generated global blueprint, $\mathcal{F}$ is the global guidance feature bank from Stage~1, $z_T^{g}$ is the initial noise, $\tilde{x}^{g}$ is the global masked image, $m^{g}$ is the global binary mask, $\mathcal{B}=\{(b_k, d_k)\}_{k=1}^{K}$ denotes the set of layout conditions (bounding boxes and object descriptions), and $c_G$ is the global prompt.
In Stage~2, $P_{\theta_2}$ generates each high-resolution patch:
\begin{equation}
    \hat{x}^{(i)} = P_{\theta_2}\left(\hat{x}^{g},\, \tilde{x}^{(i)},\, m^{(i)},\, \mathcal{F},\, p_i,\, c_L\right), \quad i=1, \dots, N_{\text{patch}},
\end{equation}
where $\hat{x}^{(i)}$ is the generated local patch for region $i$, $\tilde{x}^{(i)}$ is the local masked image, $m^{(i)}$ is the local mask, $p_i$ is the local position token encoding the coordinates of patch $i$, $c_L$ is the local prompt, and $N_{\text{patch}}$ is the total number of patches.

\subsection{Stage~1: Layout-Controllable Global Blueprint Generation}
\label{sec:stage1}
The objective of Stage~1 is to produce a low-resolution global blueprint for diffusion initialization and structural guidance features for the subsequent Stage~2.

\smallskip
\noindent \textbf{Model Architecture.} 
The backbone of Stage~1 consists of a 9-channel SD Inpainting U-Net~\cite{stable_diffusion}, which takes the masked image $\tilde{x}^g$ and a binary mask $m^g$ as additional inputs.
This allows the model to distinguish between the known source region and the target region. 
To inject layout conditions, we integrate the Layout Adapter and the Gated Fuser from InstanceDiffusion~\cite{instancediffusion} into the U-Net.
In our framework, these modules condition global blueprint generation, and the resulting blueprint is then used to guide parallel Stage~2 synthesis.
For each object $k$, the Layout Adapter generates a condition token $c_k$ by:
\begin{equation}
    c_k = \mathrm{MLP}\left([\gamma(b_k);\; e_k]\right), \quad e_k = \mathcal{T}(d_k),
\end{equation}
where $\gamma(\cdot)$, $b_k$, $d_k$, $\mathcal{T}$, and $[\,\cdot\,;\,\cdot\,]$ denote the Fourier mapping~\cite{fourier_mapping}, bounding box coordinates, object description, pre-trained CLIP text encoder~\cite{clip_score}, and concatenation, respectively.
The $K$ instance tokens are aggregated as $C = [c_1;\,\dots;\,c_K]$.
Let $H_t^g \in \mathbb{R}^{N \times d}$ denote the feature tokens from the Stage~1 U-Net at denoising step $t$.
The Gated Fuser is applied at every transformer block of the U-Net; we omit the block index $\ell$ for brevity.
The Gated Fuser injects the instance tokens $C$ into the U-Net features $H_t^g$ and produces the guidance feature $F_t^g$ as:
\begin{equation}
    F_t^g = H_t^g + \alpha_g \cdot \mathrm{Attn}\Big(Q(H_t^g), K([H_t^g; C]), V([H_t^g;C]) \Big),
\end{equation}
where $\alpha_g$ is a zero-initialized learnable gate parameter and $\mathrm{Attn}(\cdot)$ denotes standard scaled dot-product attention.
The zero-initialization preserves the pretrained generative capabilities of U-Net at the beginning of training and gradually enables conditional fusion.
At every denoising step $t$, we cache $F_t^g$ into a feature bank $\mathcal{F} = \{F_t^g\}_{t=1}^T$, which aggregates features across all Gated Fuser blocks and denoising steps, and is passed to Stage~2 as global guidance.
Since both stages share the same U-Net architecture, the guidance features are injected into the corresponding blocks of the Stage~2 U-Net with matching resolution and channel dimension.
Simultaneously, the completed denoising pass yields the clean latent $z_0^g$, which is decoded into the global blueprint $\hat{x}^g = \mathcal{D}(z_0^g)$.

\smallskip
\noindent \textbf{Attention-Guided Noise Optimization.} 
Prior to performing denoising, we optimize the initial noise $z_T^g$ to ensure semantic consistency between the source and target regions.
Specifically, by analyzing the distribution of the self-attention maps $A$ during the initial denoising step, we adjust the noise such that the target region sufficiently attends to the source region~\cite{alignnoise}.
This process proactively prevents a common issue in outpainting, where the generated region produces content that is semantically disconnected from the original image.

\smallskip
\noindent \textbf{Training Strategy.}
We train Stage~1 with standard noise-prediction objective:
\begin{equation}
    \mathcal{L}_{\mathrm{S1}} = \mathbb{E}_{z_0^{g},\, \epsilon,\, t}
    \left[\left\|
        \epsilon - \epsilon_{\theta_1}(z_t^{g},\, t;\; \tilde{x}^g,\, m^{g},\, \mathcal{B},\, c_G)
    \right\|_2^2\right],
\end{equation}
where $z_0^{g} = \mathcal{E}({x}^{g})$ is the latent of the global target image and $z_t^g$ is the noisy latent obtained by applying the forward process (Eq.~\eqref{eq:forward}) at timestep $t$.
Only the additionally integrated modules are optimized while the U-Net backbone remains frozen. 
Zero-initializing the gate parameter $\alpha_g$ ensures a stable transition to layout-aware generation.

\subsection{Stage~2: Parallel Local Patch Synthesis with Global Guidance}
\label{sec:stage2}
\begin{figure*}[t]
    \centering
    \includegraphics[width=1\linewidth]{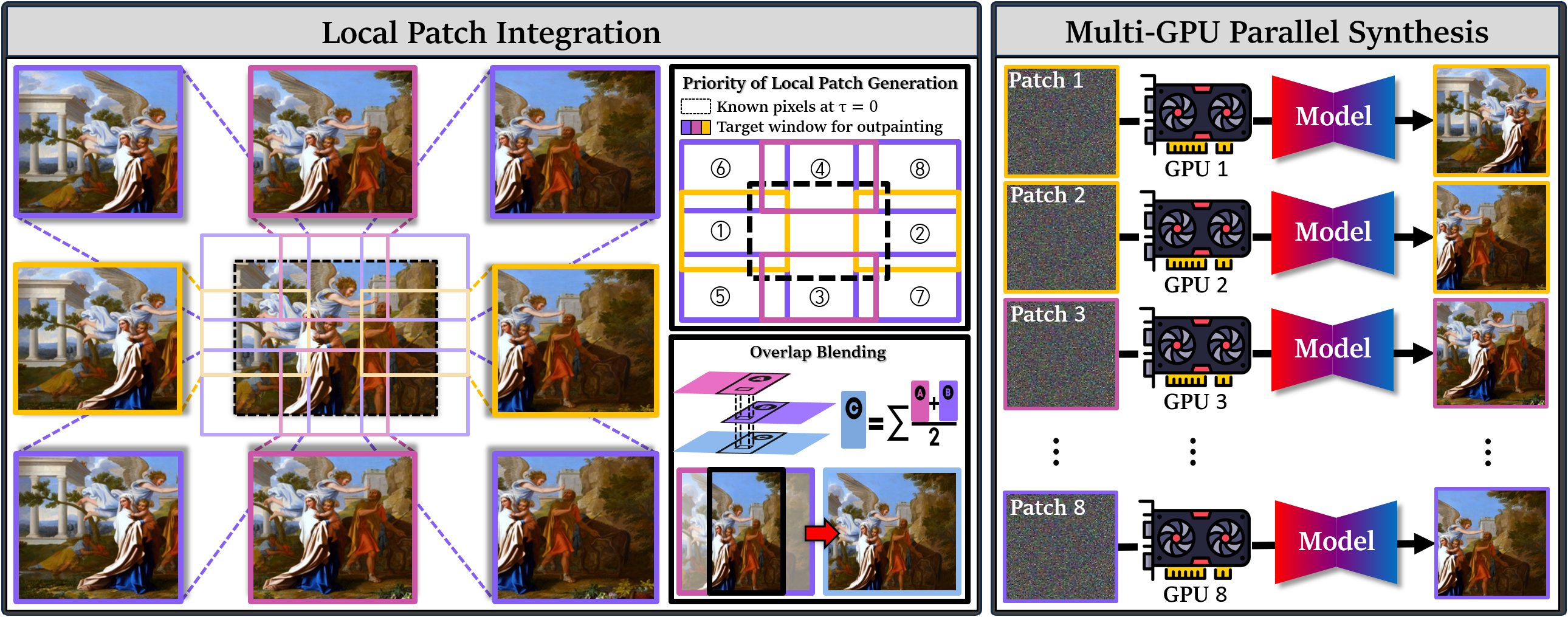}
    \caption{\textbf{Parallel Synthesis of High-resolution Local Patches.}
    \textbf{(Left)} Eight patches are arranged around the source region with overlap margins, which are blended by latent-space averaging at each denoising step for seamless composition.
    \textbf{(Right)} Each patch can be assigned to a separate GPU and denoised in parallel under global blueprint guidance.
    Patch indices follow ProOut~\cite{proout} for consistency but denote compositing priority, not sequential generation order.
    }
    \label{fig:local_patch}
\end{figure*}

The objective of Stage~2 is to generate high-resolution local patches in parallel based on the Stage~1 global blueprint.
To achieve this, Stage~2 combines three components: global guidance features from the blueprint, a patch position token that specifies the location of each patch on the full canvas, and blueprint-based initial noise construction using the low-frequency preservation property.

\smallskip
\noindent \textbf{Model Architecture.} 
The main U-Net of Stage~2 also employs the SD Inpainting U-Net~\cite{stable_diffusion} as its backbone, featuring a Gated Fuser~\cite{instancediffusion} inserted between the self-attention and cross-attention layers.
Unlike Stage~1, the position token $p_i$ in Stage~2 encodes only the spatial coordinates of each patch $i$ on the full canvas, without object description:
\begin{equation}
    p_i = \mathrm{MLP}\!\left([\gamma(b_i^{p});\; n_{\varnothing}]\right),
\end{equation}
where $b_i^{p}$ denotes the bounding box of patch $i$ on the full canvas and $n_{\varnothing}$ is a learnable null token that replaces the CLIP embedding used in Stage~1.
Let $H_t^{(i)} \in \mathbb{R}^{N \times d}$ be the main U-Net features for patch $i$, and $F_t^{g}$ the global guidance features retrieved from the Stage~1 feature bank at step $t$.
The Stage~2 Gated Fuser combines these with the local position token $p_i$:
\begin{equation}
    \tilde{H}_t^{(i)} = H_t^{(i)}+\alpha_p\cdot \mathrm{Attn}(Q(H_t^{(i)}),K(\lbrack H_t^{(i)}; F_t^g;p_i \rbrack), V(\lbrack H_t^{(i)}; F_t^g;p_i \rbrack))
\end{equation}
where $\tilde{H}_t^{(i)}$ is the fused output feature and $\alpha_p$ is the learnable gate parameter for Stage~2.
Here, $F_t^g$ provides global scene context from the blueprint, while $p_i$ specifies the location of the current patch on the full canvas, enabling blueprint-guided coordination during parallel synthesis.

\smallskip \noindent \textbf{Initial Noise Construction Based on the Global Blueprint.}
To transfer the structural information of the global blueprint to each patch, we introduce an initial noise construction strategy that leverages the low-frequency preservation property~\cite{common_diffusion_noise,freeinit}. 
First, the global blueprint $\hat{x}^{g}$ decoded from Stage~1 is upsampled to the target canvas resolution, and each patch region is cropped and re-encoded into the latent space via:
\begin{equation}
    z_{\mathrm{bp}}^{(i)} = \mathcal{E}\!\left(\mathrm{Crop}_i\!\left(
    \mathrm{Up}(\hat{x}^{g})
    \right)\right),
\end{equation}
where $z_{\mathrm{bp}}^{(i)}$ denotes the blueprint-derived latent for patch $i$, $\mathrm{Crop}_i(\cdot)$ extracts the region of patch $i$, and $\mathrm{Up}(\cdot)$ is upsampling to the target canvas resolution.
A single noise map $\epsilon^{\mathrm{full}} \sim \mathcal{N}(0,I)$ is generated at the latent resolution of the entire canvas.
Forward diffusion (Eq.~\eqref{eq:forward}) is then applied to each patch latent using the corresponding region of the global noise map:
\begin{equation}
    z_{T, \mathrm{init}}^{(i)} = \sqrt{\bar{\alpha}_T}\; z_{\mathrm{bp}}^{(i)}\; + \; \sqrt{1-\bar{\alpha}_T}\; \mathrm{Crop}_i\!\left(\epsilon^{\mathrm{full}}\right).
\end{equation}
Because a single global noise map is used, identical noise values are assigned to the same canvas locations across adjacent patches, preventing boundary discontinuities.
Due to the low-frequency preservation property, the layout and structural information of the blueprint are partially retained after forward diffusion, allowing all patches to begin denoising from structurally aligned initial states.

\smallskip \noindent \textbf{Training Strategy.} 
Similar to Stage~1, the parameters of the U-Net are frozen, and only the Gated Fuser and the Position Network that produces $p_i$ are trained.
Note that the blueprint-based initial noise construction is applied only at inference.
The Stage~2 training objective is:
\begin{equation}
    \label{stage2_loss}
    \mathcal{L}_{\mathrm{S2}} = \mathbb{E}_{i, z_0^{(i)}, \epsilon, t}
    \left[\left\|
        \epsilon - \epsilon_{\theta_2}(z_t^{(i)}, t; \tilde{x}^{(i)}, m^{(i)}, F_t^{g}, p_i, c_L)
    \right\|_2^2\right],
\end{equation}
where $c_L$ is the local prompt.
The two stages are optimized independently:
\begin{equation}
    \theta_1^{\ast} = \arg\min_{\theta_1}\, \mathcal{L}_{\mathrm{S1}},
    \qquad
    \theta_2^{\ast} = \arg\min_{\theta_2}\, \mathcal{L}_{\mathrm{S2}}.
\end{equation}
Here, only the prediction of the main U-Net contributes to Eq.~\eqref{stage2_loss}, while the frozen Stage~1 model serves solely as a feature extractor.

\subsection{Inference Pipeline}

\smallskip \noindent \textbf{Stage~1: Global Blueprint Generation.} 
Given the source image and desired canvas size, a binary mask $m^g$ is generated that designates the region outside the source as the target.
If layout conditions $\mathcal{B}$ are provided by the user, they are transformed into condition tokens $C$ via the Layout Adapter.
Before denoising, the attention-guided noise optimization described in \cref{sec:stage1} is applied to refine the initial noise.
Outpainting is then performed from the optimized noise $z_T^{g,\ast}$ to generate a low-resolution global blueprint $\hat{x}^g$.
In this process, the output features of the Gated Fuser are extracted and stored in a feature bank $\mathcal{F}$ at each denoising step, which serves as global guidance for Stage~2.

\smallskip \noindent \textbf{Stage~2: Parallel Local Patch Synthesis.} 
As shown in Figure~\ref{fig:local_patch}, in the default evaluation, the canvas is partitioned into eight local patches ($N_{\text{patch}}=8$) surrounding the source region based on the blueprint $\hat{x}^g$ generated in Stage~1.
These include four along the cardinal directions and four at the corners, following the same scheme as ProOut~\cite{proout}.
A fixed overlap margin is established between adjacent patches to prevent boundary discontinuities.
Unlike ProOut, which generates these eight patches sequentially, our method generates all patches in parallel.
Thanks to the global guidance injection and initial noise construction mechanisms described in \cref{sec:stage2}, each patch can be denoised in parallel rather than sequentially.
At each denoising step, local patches are denoised independently and then aggregated for overlap blending before being redistributed for the next step.
In multi-GPU inference, patches can be processed on different GPUs.

\smallskip \noindent \textbf{Overlap Blending.} 
For the overlap regions between adjacent patches, we blend them following~\cite{multidiffusion}. 
Specifically, at each denoising step $t$, we average the latents from all patches covering a shared spatial position $r$ on the canvas:
\begin{equation}
    \hat{z}_t(r) = \frac{\sum_{i \in \mathcal{R}(r)} z_t^{(i)}(r)}{|\mathcal{R}(r)|},
\end{equation}
where $\hat{z}_t(r)$ is the blended latent, $\mathcal{R}(r)$ the set of patches, and $z_t^{(i)}(r)$ the latent value of patch $i$.
This averaging is applied only to positions $r$ in the target region.
Positions in the source region are excluded from blending, since averaging source-region latents across patches can produce blurry results.

\smallskip \noindent \textbf{Canvas Compositing.} 
While the overlap blending ensures continuity in the latent space during denoising, the final canvas is composited in the decoded pixel space. 
Following ProOut~\cite{proout}, the eight patches are composited in a predefined order using a first-win strategy where early placed patches are preserved without being overwritten.
Finally, the unmasked source region of $\tilde{x}^g$ is pasted on top of the generated canvas, to ensure the original content remains unmodified.

\section{Experiments}
\label{sec:blind}
\subsection{Implementation Details}
\noindent \textbf{Datasets.}
We train on 16,000 HumanArt~\cite{humanart} images, 128,951 WikiArt~\cite{wikiart} images, and 413,620 LAION-5B High-Resolution~\cite{laion} images.
For evaluation, we use 6,528 IconArt~\cite{iconart} images.
Following ProOut~\cite{proout}, we apply random masking for augmentation.
Image captions are generated via BLIP~\cite{blip} for the test data and BLIP2~\cite{blip2} for the training data, with style descriptions (e.g., "a painting of") filtered out as in StyleCrafter~\cite{stylecrafter}.
The resulting captions are used as the global prompt $c_G$.
For the local prompt $c_L$ in Stage~2, we use quality-related keywords (e.g., "high quality") rather than content descriptions, as each patch covers only a partial region of the canvas.
Following InstanceDiffusion~\cite{instancediffusion}, object bounding boxes and descriptions are annotated using Grounded-SAM~\cite{sam,grounding_dino}.

\smallskip
\noindent \textbf{Evaluation Setup.} 
We compare against PQDiff~\cite{pqdiff}, SD Inpainting~\cite{stable_diffusion}, PowerPaint~\cite{powerpaint}, and ProOut~\cite{proout}.
SD Inpainting, PowerPaint, and ProOut perform progressive outpainting, and the original source image is directly pasted onto the source region of the final output.
For ProOut, we adopt the ControlNet-based variant, as no official code is available.
For PQDiff ($128 \times 128 \rightarrow 192 \times 192$), we resize and apply an SR model~\cite{drct} to match the original resolution.
We denote the variant that pastes the source back as 'copy' and the one without as 'gen'.
Following ProOut, we define the masking ratio $\rho \in (0,1)$ as the proportion of each spatial dimension to be removed from the original image.
Given an original image of size $H \times W$, we crop $\frac{\rho H}{2}$ from the top and bottom and $\frac{\rho W}{2}$ from the left and right to obtain a centered source region of size $(1{-}\rho)H \times (1{-}\rho)W$, which is then outpainted back to the original $H \times W$.
The default evaluation uses $\rho=0.333$, yielding approximately 225\% total pixels relative to the source.

\smallskip
\noindent \textbf{Training and Inference.} 
The U-Net backbones for Stage~1 and Stage~2 are initialized with pre-trained SD v1.5 Inpainting~\cite{stable_diffusion} and frozen; additional modules are initialized with pre-trained InstanceDiffusion weights and fine-tuned.
Both Stage~1 (global blueprints) and Stage~2 (local patches) operate at $512 \times 512$.
For baselines, PQDiff is trained from scratch for 150 epochs, PowerPaint for 40 epochs, and ProOut for 50 epochs; SD Inpainting uses pre-trained weights without fine-tuning.
Our Stage~1 and Stage~2 are trained for 60 and 50 epochs using AdamW~\cite{adamw} (lr=$5 \times 10^{-5}$, batch size=$512$) on two H100 and six A6000 GPUs.
Inference uses DDIM~\cite{ddim} (30 steps, CFG=3.0), and attention-guided noise optimization is applied following the default settings of AlignNoise~\cite{alignnoise}.
For all baselines, we adopt the default hyperparameters for each method.

\begin{table}[t]
    \caption{\textbf{Quantitative Results.} 
        Comparison with state-of-the-art outpainting methods across image quality, layout accuracy, and inference time. 
        Our default model achieves strong FID and layout accuracy, while Ours$^{\ast}$ further improves image quality metrics.
        Ours$^{\ast}$ denotes our method with guidance features extracted after the cross-attention layer in the Stage~1 U-Net, where the text condition is injected.
        This allows the guidance features used in Stage~2 to better reflect the text condition.
        Using 8 GPUs, our method achieves $2.4\times$ faster inference than ProOut.
        \textbf{Bold} and \underline{underline} denote the best and second-best performance, respectively.
    }
    \label{tab:tab_1}
    \centering
    \scriptsize
    \resizebox{\linewidth}{!}{%
    \renewcommand{\arraystretch}{1.2}
    \begin{tabular}{l|ccccc|cc|c}
        \toprule
        \multicolumn{1}{c|}{\multirow{2}{*}[-0.5ex]{\textbf{Methods}}} & \multicolumn{5}{c|}{\textbf{Image Quality}} & \multicolumn{2}{c|}{\textbf{Layout Acc.}} & \multicolumn{1}{c}{\textbf{Time(s)}$\downarrow$} \\
        \cmidrule(lr){2-6} \cmidrule(lr){7-8} \cmidrule(lr){9-9}
        & \textbf{FID}$\downarrow$ & \textbf{pFID}$_{256}$$\downarrow$ & \textbf{pFID}$_{512}$$\downarrow$ & \textbf{CLIP-S}$\uparrow$ & \textbf{CLIP-A}$\uparrow$ & \textbf{AP}$\uparrow$ & \textbf{IoU}$\uparrow$ & $\mu$ \\
        \midrule
        GT & -- & -- & -- & -- & -- & 0.5903 & 0.7716 & -- \\
        PQDiff (gen)~\cite{pqdiff} & 30.8942  & 54.0282  & 38.0268  & 0.1990 & 3.8542 & -- & -- & -- \\
        PQDiff (copy)~\cite{pqdiff} & 25.3032  & 41.1501  & 31.8283  & 0.2031 & 5.0144 & 0.2903 & 0.4304 & -- \\
        SD Inpainting~\cite{stable_diffusion} & 10.7598  & 9.5336   & 7.6219   & 0.2007 & 6.7041 & 0.3526 & 0.5539 & \underline{13.43} \\
        PowerPaint~\cite{powerpaint} & 10.5374  & 9.4403   & 7.4834   & 0.2003 & 6.4423 & 0.3537 & 0.5503 & 13.99 \\
        ProOut~\cite{proout} & 10.3032 & 9.2854 & 7.2515  & \underline{0.2052} & \underline{6.8585} & 0.3595 & 0.5623 & 18.31 \\
        \midrule
        \textbf{Ours}      & \underline{9.3064} & \underline{9.0635} & \underline{6.7514} & 0.2033 & 6.7949 & \underline{0.4336} & \textbf{0.6382} & \textbf{7.50} \\
        \textbf{Ours$^{\ast}$} & \textbf{9.0729} & \textbf{9.0577} & \textbf{6.6609} & \textbf{0.2055} & \textbf{6.8766} & \textbf{0.4363} & \underline{0.6361} & \textbf{7.50} \\
        \bottomrule
    \end{tabular}}
\end{table}

\begin{table}[t]
    \caption{\textbf{Quantitative Results on Expansion Ratios.} 
        Robustness comparison with state-of-the-art outpainting methods across varying expansion ratios from 200\% to 600\%, evaluated on landscape artwork images.
        Our method consistently achieves the best FID at all ratios, with competitive pFID and CLIP-based metrics, demonstrating stable visual quality even under extreme expansion.
        \textbf{Bold} denotes the best results.
    }
    \label{tab:tab_2}
    \centering
    \small
    \resizebox{\linewidth}{!}{%
    \renewcommand{\arraystretch}{1.3}
    \begin{tabular}{l|ccccc|ccccc|ccccc}
        \toprule
        \multicolumn{1}{c|}{\multirow{2}{*}{\textbf{Methods}}}& \multicolumn{5}{c|}{\textbf{200\% Ratio}} & \multicolumn{5}{c|}{\textbf{400\% Ratio}} & \multicolumn{5}{c}{\textbf{600\% Ratio}} \\
        \cmidrule(lr){2-6} \cmidrule(lr){7-11} \cmidrule(lr){12-16}
        & \textbf{FID}$\downarrow$ & \textbf{pFID}$_{256}$ & \textbf{pFID}$_{512}$ & \textbf{CLIP-S}$\uparrow$ & \textbf{CLIP-A}$\uparrow$ & \textbf{FID}$\downarrow$ & \textbf{pFID}$_{256}$ & \textbf{pFID}$_{512}$ & \textbf{CLIP-S}$\uparrow$ & \textbf{CLIP-A}$\uparrow$ & \textbf{FID}$\downarrow$ & \textbf{pFID}$_{256}$ & \textbf{pFID}$_{512}$ & \textbf{CLIP-S}$\uparrow$ & \textbf{CLIP-A}$\uparrow$ \\
        \midrule
        SD Inpainting~\cite{stable_diffusion} & 64.62 & 99.32 & 86.55 & 0.193 & 6.39 & 97.95 & 113.58 & 97.78 & 0.176 & 6.07 & 111.40 & 119.42 & 104.35 & 0.168 & 6.00 \\
        PowerPaint~\cite{powerpaint} & 52.90 & 97.41 & 85.52 & 0.200 & 6.21 & 86.36 & 104.11 & 93.74 & 0.189 & 5.42 & 101.34 & 108.00 & 99.85 & 0.183 & 4.95 \\
        ProOut~\cite{proout} & 56.22 & 97.70 & 85.66 & 0.202 & 6.64 & 81.17 & 106.21 & \textbf{91.07} & \textbf{0.202} & 6.52 & 89.84 & 110.49 & 97.66 & \textbf{0.202} & \textbf{6.41} \\
        \midrule
        \textbf{Ours} & \textbf{44.87} & \textbf{96.43} & \textbf{85.57} & \textbf{0.202} & \textbf{6.76} & \textbf{71.40} & \textbf{101.69} & 91.88 & 0.200 & \textbf{6.53} & \textbf{81.86} & \textbf{104.81} & \textbf{95.12} & 0.198 & 6.31 \\
        \bottomrule
    \end{tabular}}
\end{table}
\begin{figure}[t]
    \centering
    \includegraphics[width=1\linewidth]{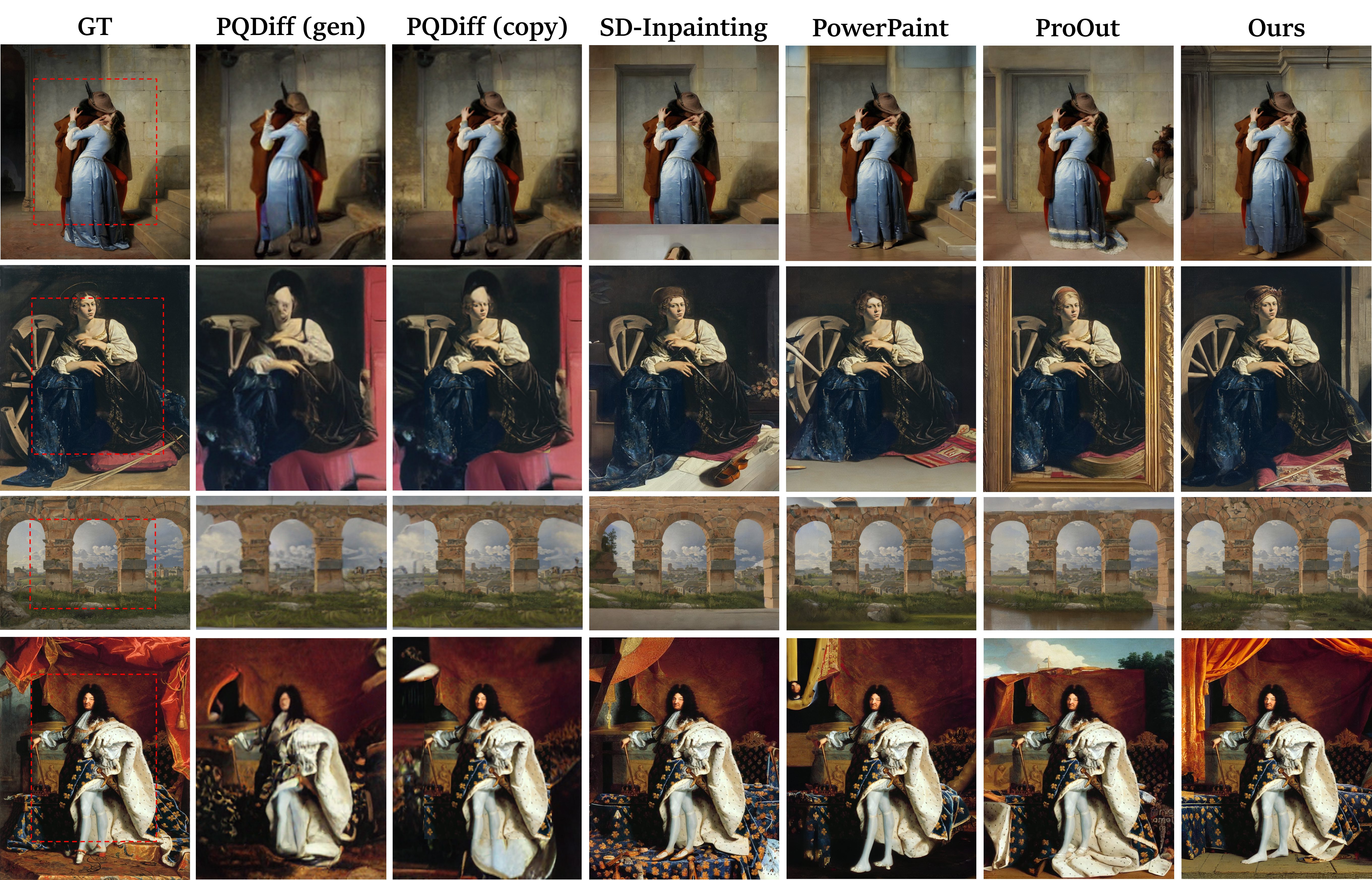}
    \caption{\textbf{Qualitative Results.}
    The red dashed box in the GT indicates the source region.
    The area outside the box is masked during inference and serves as the target region to be generated.
    PQDiff produces blurry results and SD-Inpainting creates visible seams, while PowerPaint and ProOut often introduce artificial frames or out-of-context.
    In contrast, our method synthesizes seamless, high-fidelity extensions that faithfully preserve the original style and global structure.
    }
    \label{fig:qualitative}
\end{figure}

\subsection{Comparisons}
\noindent \textbf{Quantitative Results.} 
Table~\ref{tab:tab_1} presents quantitative comparisons in terms of image quality, layout accuracy, and inference speed.
For further analysis, we additionally evaluate a variant, denoted as Ours$^{\ast}$, which extracts Stage~1 guidance features after the cross-attention layer where text conditions are injected.
This allows the guidance feature passed to Stage~2 to better reflect the text condition.
For image quality, we adopt FID~\cite{fid}, patch-FID~(pFID$_{256}$, pFID$_{512}$)~\cite{pfid_1,pfid_2}, CLIP text similarity score~(CLIP-S)~\cite{clip_score}, and CLIP aesthetic score~(CLIP-A)~\cite{laion}.
Our default model achieves strong FID and pFID scores, while Ours$^{\ast}$ further improves FID, pFID, CLIP-S, and CLIP-A.
For layout accuracy, we compute AP and IoU between the conditioned bounding boxes and those detected by Grounding DINO~\cite{grounding_dino} in the generated images, with GT detection as the upper bound.
As ours is the only framework that natively supports layout-conditioned artwork outpainting, it significantly outperforms all baselines that rely solely on text prompts.
Note that all methods benefit from pasting the source region, which contributes to the relatively moderate gap in layout accuracy scores.
For inference speed, we report the mean generation time over 100 random samples.
Using 8 GPUs, our method achieves an inference time of 7.50\,s per image, corresponding to a 2.4$\times$ reduction compared with ProOut.
Under the matched single-GPU setting, our method takes 15.73\,s per image, which is faster than ProOut at 18.31\,s but slower than SD Inpainting and PowerPaint due to the additional global planning stage and guidance injection.
Table~\ref{tab:tab_2} analyzes performance stability under increasing masking ratios from 200\% to 600\%, evaluated on 400 landscape artwork images collected from the Cleveland Museum of Art~\cite{cleveland_museum} and the Art Institute of Chicago~\cite{artic}.
The results demonstrate that our blueprint-guided strategy effectively prevents structural degradation, maintaining stable visual quality even at extreme masking ratios.
While our default evaluation uses a $3\times3$ patch grid, Stage~2 can be reconfigured to accommodate arbitrary canvas sizes.
Additional results in the supplementary material validate the generalizability of our framework across arbitrary canvas layouts~(Table~\ref{tab:supple_arbitrary_canvas}), non-artwork domains~(Table~\ref{tab:supple_general_image}), and DiT-based backbones~(Table~\ref{tab:supple_dit}).

\smallskip \noindent \textbf{Qualitative Results.}
Figure~\ref{fig:qualitative} shows the qualitative results of our method against all baselines.
We observe that existing baselines tend to generate meaningless repetitive patterns, such as frame-like textures, or produce content that lacks semantic coherence with the overall context of the source images.
Conversely, our method maintains structural and semantic consistency with the source, yielding the most visually plausible results among all compared methods.
The supplementary material further includes a qualitative comparison with Nano Banana Pro~\cite{nanobananapro} under a setting with an arbitrary expansion ratio~(Figure~\ref{fig:nanobanana}), where our method shows more reliable layout adherence.

\begin{figure}[t]
    \centering
    \includegraphics[width=1\linewidth]{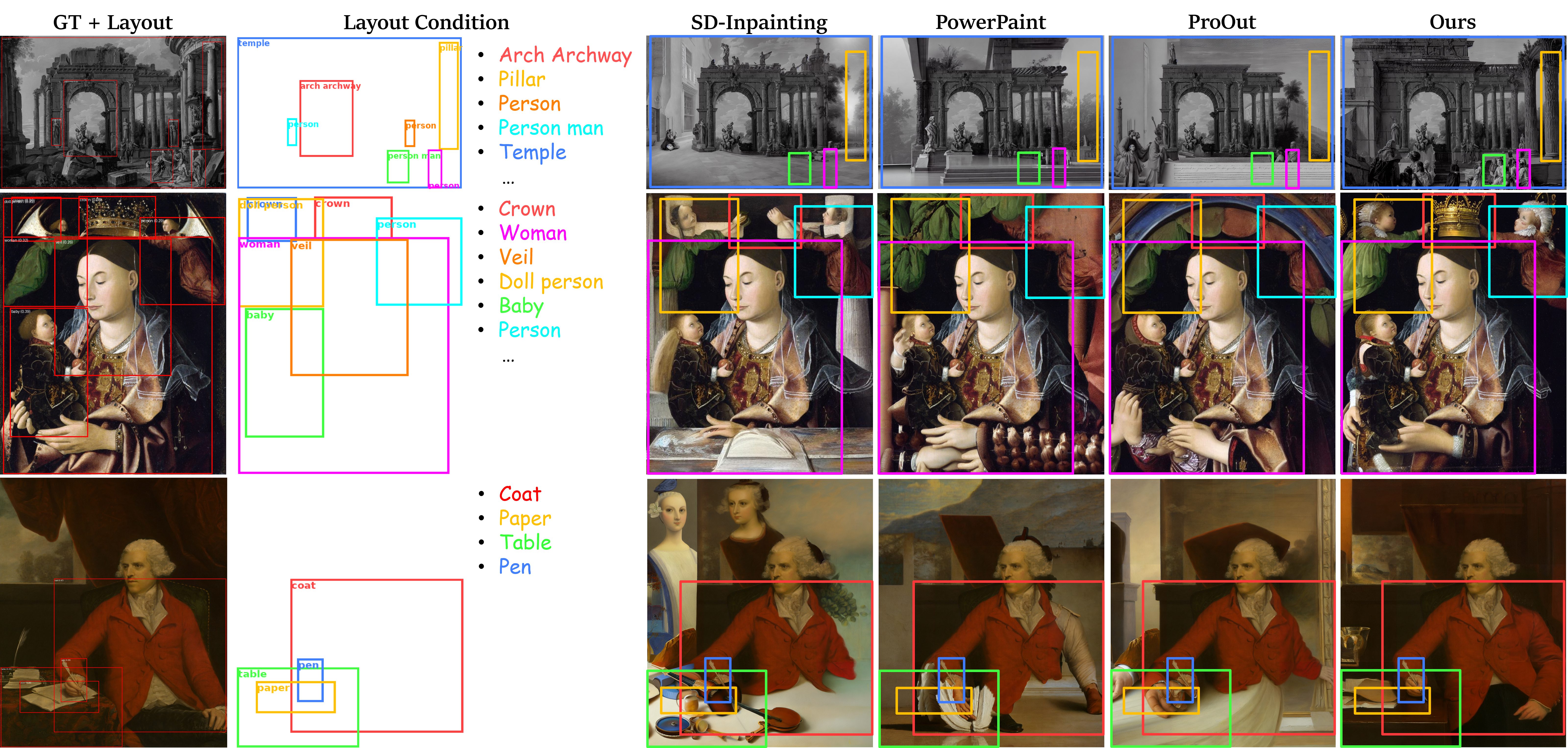}
    \caption{\textbf{Qualitative Results on Layout-guided Generation.}
    Bounding boxes and object descriptions are provided as layout conditions to guide the generation.
    Baselines rely solely on text prompts and thus struggle to control spatial placement, whereas our method accurately generates specified objects at designated locations.
    }
    \label{fig:layout}
\end{figure}

Figure~\ref{fig:layout} presents the provided layout conditions alongside the corresponding generation results.
Existing baseline models, such as SD Inpainting, PowerPaint, and ProOut, struggle with precise spatial control and often fail to generate objects at user-intended locations because they lack dedicated conditioning mechanisms beyond text prompts. 
This limitation is clearly illustrated in the second row featuring the crown, where baselines either omit the object or misplace it; in contrast, our method accurately synthesizes the object at the correct scale within the designated bounding box. 
These results confirm that text prompts alone are insufficient for fine-grained spatial control, highlighting the necessity of explicitly incorporating layout conditions for sophisticated scene composition.

\begin{table}[t]
    \caption{\textbf{Ablation Studies.} 
    We evaluate the components used to transfer the global blueprint to parallel Stage~2 synthesis: global guidance feature injection, patch position token, and forward diffusion-based initialization.
    The first row uses SD Inpainting for parallel synthesis with latent blending only, without using the global blueprint.
    Removing the guidance features or patch position tokens causes a large performance drop, while the full model achieves the best overall performance.
    }
    \label{tab:tab_3}
    \centering
    \scriptsize
    \resizebox{0.8\linewidth}{!}{%
    \renewcommand{\arraystretch}{1.2}
    \begin{tabular}{l|ccccc}
        \toprule
        \multicolumn{1}{c|}{\textbf{Methods}} & \textbf{FID}$\downarrow$ & \textbf{pFID}$_{256}\downarrow$ & \textbf{pFID}$_{512}\downarrow$ & \textbf{CLIP-S}$\uparrow$ & \textbf{CLIP-A}$\uparrow$ \\
        \midrule
        SD Inpainting (parallel)~\cite{stable_diffusion} & 37.2675  & 17.6093  & 20.3705  & 0.1932 & 5.8696 \\
        Ours (w/o Guidance Feature) & 22.2319  & 12.6436  & 11.7879  & 0.1940 & 5.8561 \\
        Ours (w/o Patch Token) & 17.9578 & 10.7008 & 8.7402 & 0.1990 & 6.2209 \\
        Ours (w/o Forward Diffusion) & 9.6048  & 9.1144  & 6.9442  & 0.2020 & 6.7598 \\
        \midrule
        \textbf{Ours} & \textbf{9.3064} & \textbf{9.0635} & \textbf{6.7514} & \textbf{0.2033} & \textbf{6.7949} \\
        \bottomrule
    \end{tabular}}
\end{table}

\subsection{Ablation Study}

\noindent We analyze the components used to transfer the global blueprint to Stage~2 synthesis: global guidance feature injection, the patch position token, and forward diffusion-based initial noise construction.
As shown in Table~\ref{tab:tab_3}, using SD Inpainting with only latent blending performs poorly, indicating that blending alone is insufficient for high-resolution outpainting.
This shows that parallel synthesis requires more than boundary smoothing, as each local patch needs to be coordinated with the Stage~1 global plan.
Removing the guidance features causes a large performance drop, showing that the global guidance features provide structural context for local patch synthesis.
Removing the patch position token also degrades performance significantly, indicating that explicit spatial encoding is necessary for coordinating each patch with the global blueprint.
Removing forward diffusion further reduces performance, showing that blueprint-based initialization provides complementary structural alignment before denoising begins.
We compare Stage~1+SR baselines in the supplementary material~(Table~\ref{tab:supple_stage1_sr}), showing that direct super-resolution of the Stage~1 blueprint does not replace Stage~2 synthesis.
Unlike cascaded diffusion pipelines~\cite{cascaded_diffusion}, our Stage~1 blueprint guides what should appear where during Stage~2 synthesis rather than serving as a low-resolution sample to be super-resolved into the final image.

\section{Conclusion}

In this paper, we introduce a global blueprint-guided two-stage diffusion framework for high-resolution artwork outpainting.
Our approach first generates a layout-conditioned, low-resolution blueprint, which then guides the parallel synthesis of high-resolution local patches.
This decoupled design effectively mitigates the error accumulation inherent in progressive methods, enables explicit spatial control via bounding boxes, and accelerates inference by up to 2.4$\times$ through multi-GPU parallel synthesis.
Extensive experiments on large-scale artwork datasets confirm that our framework achieves superior visual fidelity and structural consistency, while offering enhanced spatial controllability and inference efficiency compared to existing baselines.
These results demonstrate that global blueprint-guided planning offers a promising paradigm for scalable and controllable high-resolution artwork outpainting.

\section*{Acknowledgements}

This work was partly supported by Institute of Information \& communications Technology Planning \& Evaluation (IITP) grant funded by the Korea government (MSIT) (No. RS-2025-25422680, Metacognitive AGI Framework and its Applications, 25\%), Institute of Information \& communications Technology Planning \& Evaluation (IITP) grant funded by the Korea government (MSIT) (No. RS-2020-II201373, Artificial Intelligence Graduate School Program (Hanyang University), 25\%), Institute of Information \& communications Technology Planning \& Evaluation (IITP) under the Leading Generative AI Human Resources Development grant funded by the Korea government (MSIT) (No. IITP-2026-RS-2026-25544647, 25\%), and the AI Computing Infrastructure Enhancement (GPU Rental Support) User Support Program funded by the Ministry of Science and ICT (MSIT), Republic of Korea (No. RQT-25-120160, 25\%).

\bibliographystyle{splncs04}
\bibliography{main}

\clearpage

\setcounter{section}{0}
\setcounter{figure}{0}
\setcounter{table}{0}
\setcounter{equation}{0}

\renewcommand{\thesection}{S\arabic{section}}
\renewcommand{\thefigure}{S\arabic{figure}}
\renewcommand{\thetable}{S\arabic{table}}
\renewcommand{\theequation}{S\arabic{equation}}

\renewcommand{\theHsection}{suppl.\arabic{section}}
\renewcommand{\theHsubsection}{suppl.\arabic{section}.\arabic{subsection}}
\renewcommand{\theHfigure}{suppl.\arabic{figure}}
\renewcommand{\theHtable}{suppl.\arabic{table}}
\renewcommand{\theHequation}{suppl.\arabic{equation}}

\title{Supplementary Material for ``High-Resolution Artwork Outpainting with Global Blueprint Guidance and Layout Control''} 

\titlerunning{Supplementary Material}

\author{Junha Kim\orcidlink{0009-0003-0939-0113} \and
Hyunjoon Park\orcidlink{0009-0008-1860-3172} \and
Donghyeon Cho\textsuperscript{\ensuremath{\dagger}}\orcidlink{0000-0002-2184-921X}}

\authorrunning{J.~Kim et al.}
\institute{Hanyang University, Seoul, Republic of Korea\\
\email{\{poohoh,junippini83,doncho\}@hanyang.ac.kr}}

\maketitle

\begingroup
\renewcommand{\thefootnote}{\ensuremath{\dagger}}
\footnotetext{Corresponding author.}
\endgroup

\section{Overview}
The Supplementary Materials expand the condensed implementation details of the main paper and provide the additional analyses supporting the main experimental claims.
It is organized into four parts: Additional Implementation Details (\cref{sec:supple_implement}) on architecture and evaluation details, Additional Analysis (\cref{sec:supple_analysis}) on ablations, sensitivity studies, and controlled comparisons, Extended Baseline Comparison (\cref{sec:supple_baseline}) on robustness to baseline settings, and Qualitative Results (\cref{sec:supple_qualitative}) on additional visual comparisons and failure cases.

\begin{figure}
    \centering
    \includegraphics[width=0.9\linewidth]{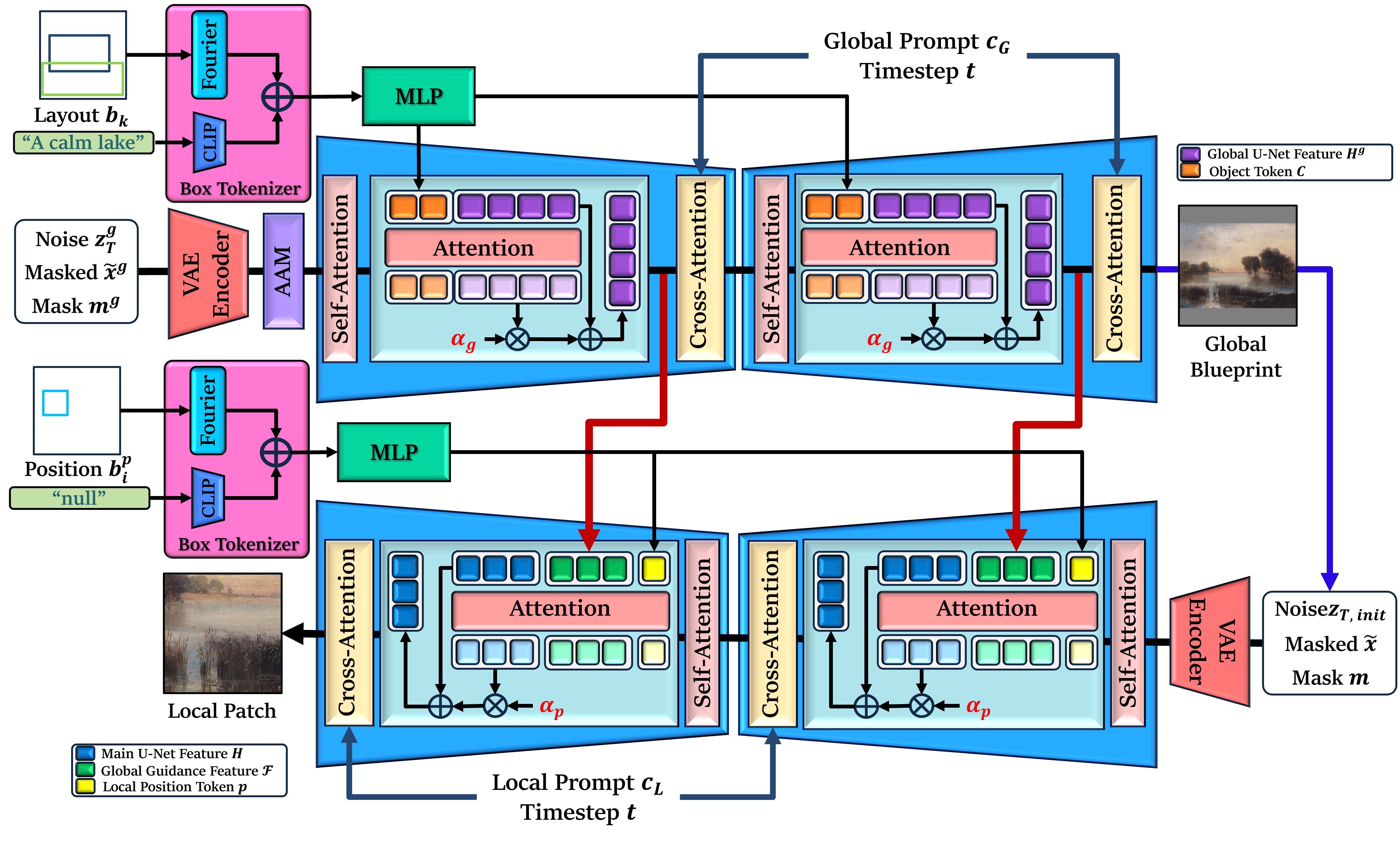}
    \caption{\textbf{Detailed Architectural Overview and Data Flow.}
        This diagram expands the condensed pipeline in the main paper, showing the complete input-output connections of each module.
        Stage~1 processes layout and text data to generate a global blueprint while exporting guidance features $\mathcal{F}$ (red arrows).
        Stage~2 initializes each local patch from the blueprint via forward diffusion (blue arrow) and denoises all patches in parallel under $\mathcal{F}$ and the position token $p_i$.
    }
    \label{fig:pipeline_detail}
\end{figure}

\section{Additional Implementation Details}
\label{sec:supple_implement}

\subsection{Model Structure}
This section provides architectural details that are abbreviated in the main text.
Fig.~\ref{fig:pipeline_detail} illustrates the data flow of our two-stage framework, showing how Stage~1 exports guidance features $\mathcal{F}$ and generates the blueprint $\hat{x}^g$, and how each local patch in Stage~2 is initialized via forward diffusion and denoised in parallel.

\begin{figure}
    \centering
    \includegraphics[width=0.7\linewidth]{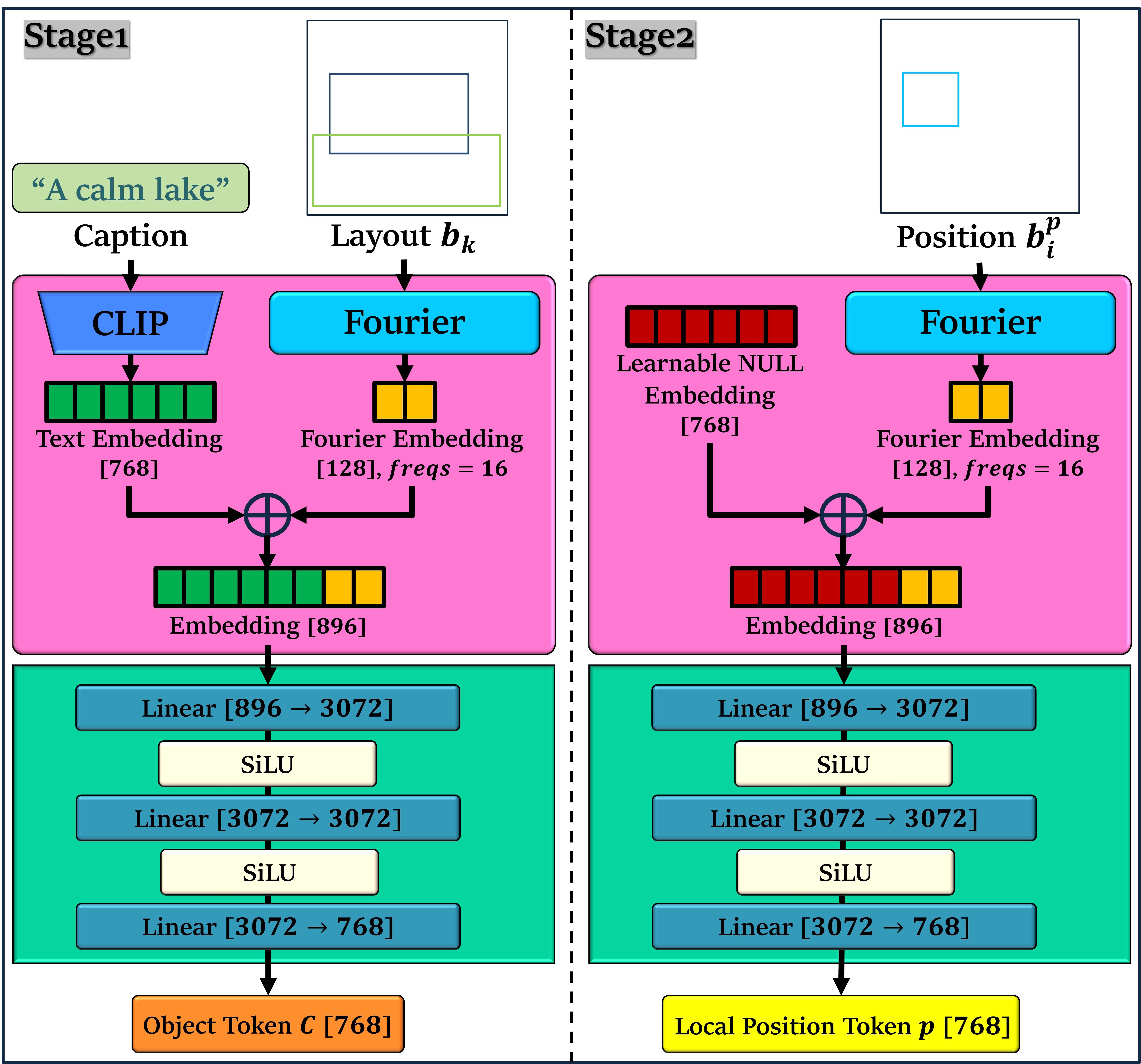}
    \caption{\textbf{Architecture of Layout Adapter and Position Network.}
        The two modules share the same architecture but are used for different conditioning signals in the two stages.
        In Stage~1, the Layout Adapter combines the CLIP text embedding of each object description with the Fourier-encoded bounding box $b_k$ to produce an object token $c_k$.
        In Stage~2, the Position Network replaces the text embedding with a learnable null embedding and combines it with the Fourier-encoded patch position box $b_i^p$ to produce the local position token $p_i$.
    }
    \label{fig:layout_adapter}
\end{figure}

\smallskip \noindent \textbf{Layout Adapter and Position Network.}
\noindent As shown in Fig.~\ref{fig:layout_adapter}, the Layout Adapter in Stage~1 and Position Network in Stage~2 share the same architecture but serve different purposes.
The Layout Adapter encodes object-level semantic and spatial conditions by combining the Fourier-encoded object box $\gamma(b_k)$ with the CLIP text embedding~\cite{clip_score} $e_k$ of the object description to produce the object token $c_k$.
The Position Network instead encodes patch-level spatial conditions by replacing the text embedding with a learnable null token $n_{\varnothing}$ and mapping the patch position box $b_i^p$ to the local position token $p_i$.

\begin{figure}
    \centering
    \includegraphics[width=0.7\linewidth]{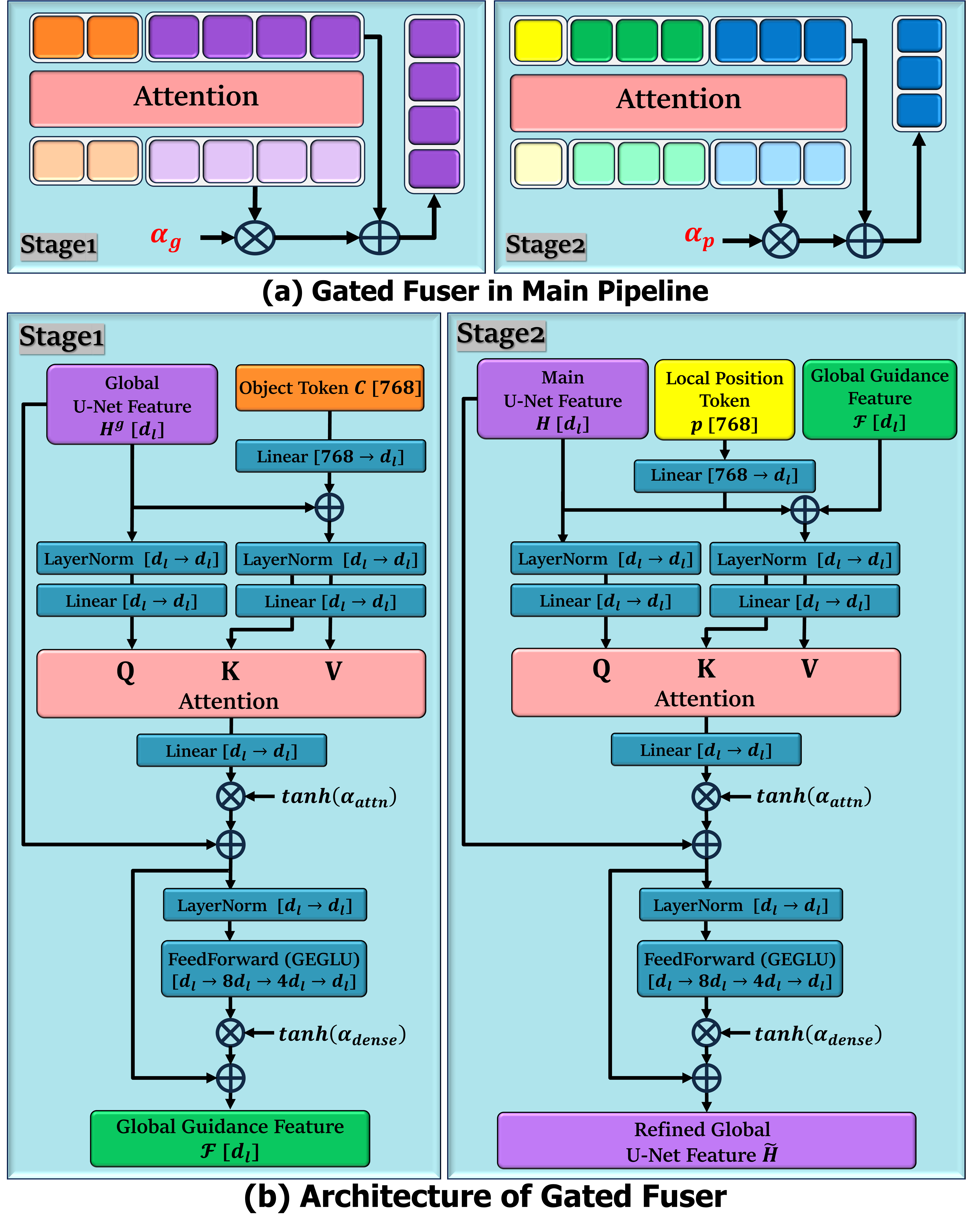}
    \caption{\textbf{Architecture of the Gated Fuser.}
        \textbf{(a) Pipeline integration:} Overview of Gated Fuser modules embedded within the two-stage framework.
        \textbf{(b) Detailed design:} Internal structure of the modules for each stage. 
        The primary distinction lies in their inputs.
        Stage 1 processes U-Net features and object tokens, whereas Stage 2 incorporates local position token $p_i$ and global guidance features for refined synthesis.
    }
    \label{fig:gated_fuser}
\end{figure}

\smallskip \noindent \textbf{Gated Fuser.}
Fig.~\ref{fig:gated_fuser} details the structure of the Gated Fuser.
In both stages, the query is projected from the U-Net features, while the key and value are projected from the concatenation of the U-Net features and auxiliary condition tokens.
In Stage~1, the auxiliary tokens are the object tokens $C$ after a linear projection to match the U-Net feature dimension.
In Stage~2, the global guidance features $F_t^g$ and the position token $p_i$ are also concatenated.
The output of the attention is scaled by a zero-initialized gate $\mathrm{tanh}(\alpha_{attn})$ and added as a residual, then passed through a LayerNorm and feed-forward block gated by $\mathrm{tanh}(\alpha_{dense})$.

\subsection{Evaluation Setup}
\noindent \textbf{Evaluation Setting.}
The main quantitative results in the main paper are reported under an evaluation setting in which our method receives additional layout conditions derived from the full target image, together with an image caption.
Specifically, at inference time, we provide our model with bounding boxes and object descriptions inferred from the target image in addition to the image caption, whereas the compared baselines rely on text prompts only.
Because both the image caption and the layout annotations are obtained from the full image, the main tables should be interpreted as a conditional comparison with additional structural priors, not as a strictly source-only comparison.
Experiments without layout conditions are presented separately in \cref{sec:supple_analysis}.

\smallskip \noindent \textbf{Inference-Time Inputs.}
Under the main evaluation setting, our method receives image caption $c_G$ together with bounding boxes and object descriptions.
ProOut~\cite{proout} receives the same image caption $c_G$ but no layout condition.
PowerPaint~\cite{powerpaint} does not use the image caption $c_G$ as a condition in our reported setting.
Instead, it follows the default contextual prompting scheme of PowerPaint, using the learnable prompt embeddings $P_{ctxt}$ and $P_{obj}$.
For each patch, the conditional prompt prepends the contextual prompt $P_{ctxt}$ together with the phrase ``empty scene blur'' to the local quality prompt $c_L$, while the unconditional prompt prepends the object prompt $P_{obj}$ to the negative prompt.
SD Inpainting~\cite{stable_diffusion} does not use the image caption $c_G$ or any layout condition at inference and instead uses only the fixed local quality prompt $c_L$ and the fixed negative prompt.
PQDiff~\cite{pqdiff} receives neither text nor layout conditions and uses only the centered source crop.
In addition, the original centered source region is pasted back into the final composite for all methods except PQDiff~(\emph{gen}).

\begin{figure}
    \centering
    \includegraphics[width=0.5\linewidth]{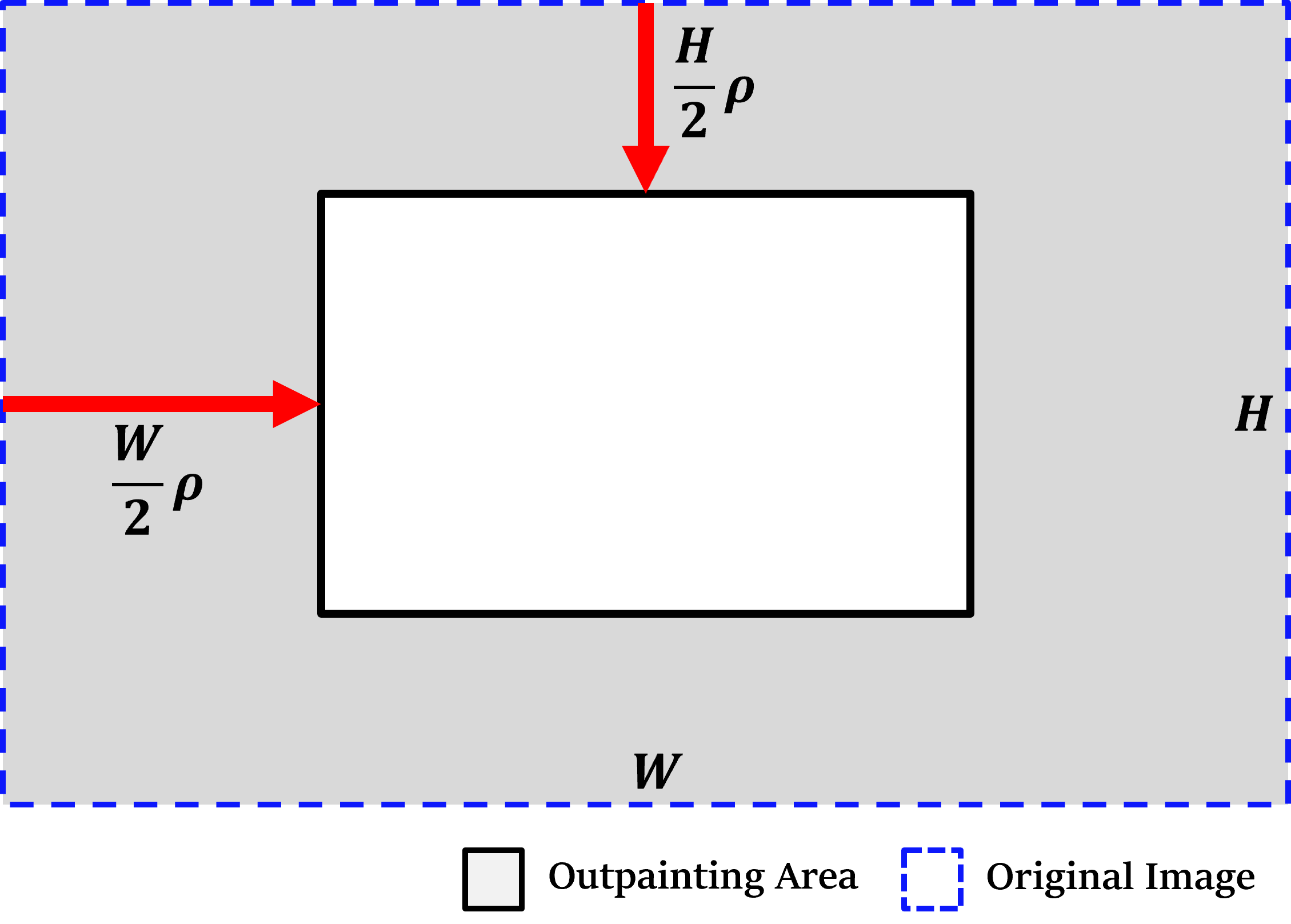}
    \caption{\textbf{Definition of the Outpainting Region.} 
    The masking ratio $\rho$ determines the size of the centered source region within an image of size ($H\times W$).
    The offsets $\frac{\rho W}{2}$ and $\frac{\rho H}{2}$ define the boundary between the source and the generated outer area.
    }
    \label{fig:overlap}
\end{figure}

\smallskip \noindent \textbf{Masking Ratio and Source Preservation.}
For inference, following ProOut~\cite{proout}, we define the masking ratio $\rho \in (0,1)$ as the proportion removed from each spatial dimension.
Given an original image of size $H \times W$, Fig.~\ref{fig:overlap} illustrates how the masking ratio $\rho$ determines the centered source region by removing $\frac{H}{2}\rho$ from the top and bottom and $\frac{W}{2}\rho$ from the left and right; the remaining outer area then forms the outpainting target region.
The default evaluation uses $\rho = 0.333$, yielding approximately $225\%$ total pixels relative to the source.
Before cropping, images whose long edge exceeds 1408 pixels are downsampled to a long edge of 1408 while preserving aspect ratio.
For the progressive diffusion methods, the original centered source crop is pasted back into the final composite so that the original source region is preserved exactly.
PQDiff~\cite{pqdiff} is reported in both a \emph{copy} variant, where the source is pasted back, and a \emph{gen} variant, where the model output is retained everywhere.

\subsection{Dataset and Annotation}
\noindent \textbf{Caption and Layout Preparation.}
Building on the training and evaluation datasets introduced in \cref{sec:supple_implement}, we next describe how captions and layout annotations are prepared.
For each image, we prepare a global caption $c_G$, instance-level bounding boxes, and object descriptions.
For the training data, captions $c_G$ are generated with BLIP2~\cite{blip2}, whereas for the test data they are generated with BLIP~\cite{blip}.
As in the main paper, generic style prefixes such as ``a painting of'' are removed from the captions.
Bounding boxes and object descriptions are obtained with Grounded-SAM~\cite{sam,grounding_dino}.
Following ProOut~\cite{proout}, we use fixed quality-oriented prompts for local and negative conditioning.
The local prompt $c_L$ is therefore not object-specific and instead uses ``harmonized painting, high resolution, best quality, high quality, harmonized simple background'', while the negative prompt is fixed to ``ugly, nsfw, worst quality, watermark, signature, logo''.

\smallskip \noindent \textbf{Data Augmentation.}
Following ProOut~\cite{proout}, we use the same random-masking augmentation pipeline for training each stage, producing the stage-specific masked inputs and masks required for global and local generation.
PQDiff is the only trained baseline that retains its original preprocessing procedure instead of using this augmentation pipeline.

\subsection{Training Configuration}
\noindent \textbf{Our Model.}
Both stages are initialized from the pretrained Stable Diffusion v1.5 Inpainting~\cite{stable_diffusion} checkpoint, while the additional conditioning modules are initialized from pretrained InstanceDiffusion~\cite{instancediffusion} weights before fine-tuning.
In both stages, the pretrained Inpainting U-Net backbone remains frozen.
Stage~1 is trained for 60 epochs and Stage~2 for 50 epochs using AdamW~\cite{adamw} with learning rate $5\times10^{-5}$.
The global batch size is 480 for Stage~1 and 520 for Stage~2, with 5,000 warmup steps for both stages.
During Stage~1 training, box-conditioned inputs are randomly dropped with probability 0.1.
During Stage~2 training, the local text prompt is dropped with probability 0.1 while the frozen Stage~1 blueprint pathway remains active.

\smallskip \noindent \textbf{Baselines.}
For the main comparison, PQDiff~\cite{pqdiff} is trained from scratch for 150 epochs and evaluated with the EMA checkpoint.
Our ControlNet-based ProOut~\cite{proout} reproduction is trained for 50 epochs following the setup proposed in ProOut.
Powerpaint~\cite{powerpaint} is trained in its Stable Diffusion v1.5 variant, and the main paper reports the epoch-40 checkpoint.
SD Inpainting~\cite{stable_diffusion} uses the released Stable Diffusion v1.5 Inpainting checkpoint without additional fine-tuning.

\subsection{Default Inference Settings}
\noindent \textbf{Our Two-Stage Model.}
Unless otherwise stated, both stages use DDIM~\cite{ddim} sampling with 30 steps, $\eta=0$, seed 42, classifier-free guidance scale 3.0, and the fixed negative prompt defined above.
Before Stage~1 denoising, we apply the AlignNoise~\cite{alignnoise} AAM procedure, which optimizes the initial noise using attention maps, with the default setting of 10 iterations and learning rate $10^{-2}$.
Stage~1 then generates the global blueprint $\hat{x}^g$ and exports the timestep-wise guidance features $\mathcal{F}$.
In Stage~2, the blueprint $\hat{x}^g$ is upsampled to the target canvas resolution, and each patch region is then cropped and re-encoded into latent space.
Before denoising, the resulting local latent is forward-diffused to timestep $T$ using the corresponding region of a shared global noise map.
The eight surrounding patches are then denoised in parallel under the global guidance features $\mathcal{F}$ and their position tokens.

\smallskip \noindent \textbf{Baselines.}
ProOut~\cite{proout}, PowerPaint~\cite{powerpaint}, and SD Inpainting~\cite{stable_diffusion} share a common progressive configuration on IconArt, using $512\times512$ patches in the order $N \rightarrow E \rightarrow S \rightarrow W \rightarrow NE \rightarrow NW \rightarrow SE \rightarrow SW$, DDIM with 30 steps, $\eta=0$, seed 42, and classifier-free guidance scale 7.0.
Here, $N$, $E$, $S$, and $W$ denote north, east, south, and west.
PQDiff~\cite{pqdiff} uses its native $128\times128 \rightarrow 192\times192$ generation setting with DPM-Solver-fast sampling, after which the output is restored to the target aspect ratio and refined with DRCT~\cite{drct} $\times4$ super-resolution.

\subsection{Baseline Reproduction Details}
\noindent \textbf{ProOut.}
No official training code is publicly available for ProOut~\cite{proout}.
We therefore implement an unofficial ControlNet-based reproduction using the official ControlNet~\cite{controlnet} codebase, initialized from Stable Diffusion v1.5 Inpainting~\cite{stable_diffusion}.
The image caption is used as the global prompt $c_G$ for ControlNet, while the U-Net uses a local prompt following the shared quality-prompt template.
On IconArt~\cite{iconart}, ProOut performs sequential progressive outpainting using the image caption as the global prompt $c_G$.

\smallskip \noindent \textbf{PowerPaint.}
We train PowerPaint~\cite{powerpaint} with two learnable prompt embeddings for context and object conditioning, denoted by $P_{ctxt}$ and $P_{obj}$, respectively.
The shape prompt $P_{shape}$ is omitted because our datasets do not contain segmentation masks, and each remaining embedding is composed of 10 vectors.
During training, the contextual and object prompts are used with equal probability, with prompt dropout set to 0.1 for both.

\smallskip \noindent \textbf{PQDiff.}
PQDiff~\cite{pqdiff} is trained and evaluated at $128\times128 \rightarrow 192\times192$ resolution and does not use text conditioning at inference.
For a target canvas of size $W\times H$, we first generate a square $192\times192$ output from the centered $128\times128$ source image.
We then resize the result bicubically to $(\lceil W/4 \rceil, \lceil H/4 \rceil)$ and apply DRCT $\times4$ super-resolution to restore the original scale.

\smallskip \noindent \textbf{SD Inpainting.}
SD Inpainting~\cite{stable_diffusion} is used as a baseline without fine-tuning.

\subsection{Overlap Planning and Canvas Composition}
\noindent \textbf{Overlap Planning.}
All progressive methods use a centered $3\times3$ canvas layout with patch size $P=512$ and eight generated patches placed around the source region.
If the horizontal and vertical overlaps between neighboring patches are $o_x$ and $o_y$, the full canvas satisfies $W = P + 2(P-o_x)$ and $H = P + 2(P-o_y)$, which yields the nominal overlaps $o_x^\ast = \frac{3P-W}{2}$ and $o_y^\ast = \frac{3P-H}{2}$.
In practice, these values are adjusted and clamped to a valid range.

\smallskip \noindent \textbf{Blending and Final Composition.}
During Stage~2, the eight patches are denoised in parallel.
Overlapping generated regions are averaged in latent space at each denoising step, after which the decoded patches are composed on the canvas in a fixed priority order.
The original unmasked source region is finally pasted back onto the canvas.

\subsection{Evaluation Metrics and Considerations}
\noindent \textbf{Metrics.}
We compute FID~\cite{fid} and pFID~\cite{pfid_1,pfid_2} using \texttt{torch-fidelity}.
CLIP-S~\cite{clip_score} and CLIP-A are computed with OpenCLIP ViT-L/14.
For pFID, we use patch sizes 256 and 512, with a fixed seed of 42 for patch cropping.
Layout accuracy is measured by AP and IoU between the conditioning boxes and the boxes detected by Grounding DINO~\cite{grounding_dino} in generated images.

\smallskip \noindent \textbf{Interpretation of the Main Results.}
The main quantitative tables evaluate our method under a layout-conditioned setting and therefore show how the full method performs relative to the baselines when additional structural guidance is available.
This comparison should not be interpreted as a strictly matched source-only setting.
To isolate the effect of the layout condition itself, we additionally report companion experiments without layout conditions in \cref{sec:supple_analysis}.
As shown there, our method remains competitive even without layout conditioning, indicating that the gains in the main tables are not solely due to the additional layout input.
This suggests that the layout condition is introduced primarily for controllability rather than merely to boost quantitative performance.

\smallskip \noindent \textbf{Experimental Limitations.}
The annotation pipeline uses Grounded-SAM, whereas the layout metric uses Grounding DINO, which belongs to the same detector family.
We therefore interpret the AP and IoU values as detector-dependent and use them for relative comparison.
We note that no practical layout-conditioned GAN baseline was available for reproduction in our setting.

\section{Additional Analysis}
\label{sec:supple_analysis}
\begin{table}[t]
    \caption{\textbf{Layout Condition Ablation.}
        Removing layout annotations slightly degrades image-quality metrics, while explicit layout conditioning yields clearer gains in spatial controllability.
        Stage~1 denotes low-resolution blueprint generation, and Stage~2 denotes the final high-resolution outpainting results.
    }
    \label{tab:supple_layout_ablation}
    \centering
    \scriptsize
    \resizebox{0.8\linewidth}{!}{%
    \renewcommand{\arraystretch}{1.2}
    \begin{tabular}{l|ccccc|cc}
        \toprule
        \multicolumn{1}{c|}{\multirow{2}{*}{\textbf{Methods}}} & \multicolumn{5}{c|}{\textbf{Image
        Quality}} & \multicolumn{2}{c}{\textbf{Layout Acc.}} \\
        \cmidrule(lr){2-6} \cmidrule(lr){7-8}
        & \textbf{FID}$\downarrow$ & \textbf{pFID}$_{256}$$\downarrow$ & \textbf{pFID}$_{512}$$
    \downarrow$ & \textbf{CLIP-S}$\uparrow$ & \textbf{CLIP-A}$\uparrow$ & \textbf{AP}$\uparrow$ &
    \textbf{IoU}$\uparrow$ \\
        \midrule
        Ours w/o layout (Stage 1) & 12.0213 & -- & -- & 0.2057 & 6.4046 & 0.3134 & 0.4515 \\
        Ours w/ layout (Stage 1)  & 10.9963 & -- & -- & 0.2072 & 6.4663 & 0.4550 & 0.5875 \\
        Ours w/o layout (Stage 2) & 9.7111 & 9.0660 & 6.7552 & 0.2029 & 6.7906 & 0.3973 & 0.5995 \\
        Ours w/ layout (Stage 2)  & 9.3064 & 9.0635 & 6.7514 & 0.2033 & 6.7949 & 0.4336 & 0.6382 \\
        \bottomrule
    \end{tabular}}
\end{table}
Table~\ref{tab:supple_layout_ablation} isolates the effect of the layout condition in our two-stage framework.
Stage~1 corresponds to low-resolution blueprint generation, and Stage~2 corresponds to the final high-resolution results.
This shows that layout conditioning mainly contributes controllability rather than being the sole source of image-quality improvement.
We omit pFID for Stage~1 because many blueprint outputs are too small for valid $256\times 256$ or $512\times512$ patch cropping.

\begin{table}
    \caption{\textbf{Stage~1+SR Comparison.}
        We test whether Stage~2 can be replaced by direct super-resolution of the Stage~1 blueprint using off-the-shelf SR models, DRCT-L $\times4$ and PiSA-SR $\times4$.
        The Stage~1+SR methods show substantially worse FID and pFID than the full two-stage model, indicating that the blueprint serves as a structural guide for Stage~2 rather than a low-resolution target for super-resolution.
    }
    \label{tab:supple_stage1_sr}
    \centering
    \scriptsize
    \resizebox{0.8\linewidth}{!}{%
    \renewcommand{\arraystretch}{1.2}
    \begin{tabular}{l|ccccc}
        \toprule
        \multicolumn{1}{c|}{\textbf{Methods}} & \textbf{FID}$\downarrow$ & \textbf{pFID}$_{256}$$
        \downarrow$ & \textbf{pFID}$_{512}$$\downarrow$ & \textbf{CLIP-S}$\uparrow$ & \textbf{CLIP-A}$
        \uparrow$ \\
        \midrule
        Stage~1 + DRCT-L~\cite{drct} & 22.0840 & 42.0694 & 31.2430 & 0.2047 & 5.8405 \\
        Stage~1 + PiSA-SR~\cite{pisa_sr} & 36.4412 & 41.9521 & 36.0056 & 0.2042 & 6.5461 \\
        \midrule
        Ours (w/o layout) & 9.7111 & 9.0660 & 6.7552 & 0.2029 & 6.7906 \\
        Ours (w/ layout)  & 9.3064 & 9.0635 & 6.7514 & 0.2033 & 6.7949 \\
        \bottomrule
    \end{tabular}}
\end{table}

Table~\ref{tab:supple_stage1_sr} evaluates whether SR models can replace Stage~2 by super-resolving the Stage~1 blueprint.
For the Stage~1+SR methods, we directly super-resolve the Stage~1 output without pasting the known source region, since pasting the source before or after SR produced visible seams.
Preserving the original source region is particularly important in artwork outpainting, but directly pasting it into an SR output creates visible discontinuities at the boundary.
Without pasting, however, the source region deviates from the original input.
This reflects a practical limitation of treating the blueprint as an SR target in our setting.
Although the Stage~1 blueprint captures the global composition, directly applying SR to this output performs substantially worse than our Stage~2 synthesis, especially in FID and pFID.
This confirms that, unlike cascaded diffusion pipelines~\cite{cascaded_diffusion}, the Stage~1 blueprint is not a low-resolution sample to be super-resolved.
Instead, it provides structural guidance for Stage~2 through guidance features and blueprint-based initialization.

\begin{table}
    \caption{\textbf{Arbitrary Canvas Evaluation.}
    We evaluate our method without retraining under wide, tall, and large-canvas settings.
    Stage~2 flexibly reconfigures the patch grid and overlap according to the target canvas size and aspect ratio.
    Our method consistently improves FID over progressive baselines across all settings.
    }
    \label{tab:supple_arbitrary_canvas}
    \centering
    \scriptsize
    \resizebox{\linewidth}{!}{%
    \renewcommand{\arraystretch}{1.15}
    \begin{tabular}{c l|ccccc|c}
        \toprule
        \textbf{Setting} & \multicolumn{1}{c|}{\textbf{Methods}} & \textbf{FID}$\downarrow$ & \textbf{pFID}$_{256}\downarrow$ & \textbf{pFID}$_{512}\downarrow$ & \textbf{CLIP-S}$\uparrow$ & \textbf{CLIP-A}$\uparrow$ & \textbf{Time}$\downarrow$ \\
        \midrule
        \multirow{5}{*}{Wide} & SD Inpainting~\cite{stable_diffusion} & 96.00 & 114.62 & 105.82 & 0.1875 & 5.7728 & \textbf{36.26} \\
        & PowerPaint~\cite{powerpaint} & 79.44 & 107.21 & 102.56 & 0.1939 & 5.5584 & \underline{37.16} \\
        & ProOut~\cite{proout} & 78.12 & 106.10 & 106.65 & \textbf{0.1973} & 5.9476 & 48.80 \\
        & Ours (w/o layout) & \underline{58.40} & \textbf{100.28} & \textbf{99.73} & 0.1937 & \underline{6.0083} & 38.11 \\
        & \textbf{Ours (w/ layout)} & \textbf{54.08} & \underline{103.36} & \underline{100.48} & \underline{0.1947} & \textbf{6.0220} & 38.04 \\
        \midrule
        \multirow{5}{*}{Tall} & SD Inpainting~\cite{stable_diffusion} & 103.96 & 115.74 & 105.60 & 0.1828 & 5.7610 & \textbf{29.86} \\
        & PowerPaint~\cite{powerpaint} & 92.69 & 103.82 & 100.98 & 0.1942 & 5.6202 & \underline{30.55} \\
        & ProOut~\cite{proout} & 89.18 & 108.25 & 100.28 & \textbf{0.1952} & 5.9261 & 40.29 \\
        & Ours (w/o layout) & 66.57 & \underline{98.98} & \underline{96.38} & 0.1942 & \textbf{6.0968} & 32.09 \\
        & \textbf{Ours (w/ layout)} & \textbf{61.44} & \textbf{98.42} & \textbf{96.35} & \textbf{0.1952} & \underline{6.0502} & 31.89 \\
        \midrule
        \multirow{5}{*}{Large} & SD Inpainting~\cite{stable_diffusion} & 130.55 & 122.59 & 126.88 & 0.1718 & 5.9854 & \textbf{63.69} \\
        & PowerPaint~\cite{powerpaint} & 96.22 & 109.11 & 114.32 & 0.1892 & 5.5375 & 66.67 \\
        & ProOut~\cite{proout} & 95.93 & 112.21 & 122.35 & 0.1942 & 6.4052 & 85.20 \\
        & Ours (w/o layout) & \underline{63.55} & \underline{104.28} & \textbf{112.26} & \underline{0.1995} & \underline{6.8197} & \underline{64.21} \\
        & \textbf{Ours (w/ layout)} & \textbf{61.20} & \textbf{101.08} & \underline{113.43} & \textbf{0.2000} & \textbf{6.8525} & 64.23 \\
        \bottomrule
    \end{tabular}}
\end{table}

Table~\ref{tab:supple_arbitrary_canvas} evaluates our method under arbitrary canvas settings without retraining.
Each setting contains 400 high-resolution artwork images collected from public-domain museum sources.
The large-canvas setting uses images whose short side is at least 2560 pixels, resulting in 42.23 generated patches per image on average.
Across wide, tall, and large-canvas settings, our method consistently improves FID over progressive baselines.

\begin{table}
    \caption{\textbf{General-Image Evaluation.}
    We evaluate our method on Open Images across natural, indoor, and human-centric scenes without retraining.
    Each category contains 400 images and follows the same expansion setting as the main evaluation.
    The results show that our method generalizes beyond artwork images while remaining competitive in CLIP-based metrics.
    }
    \label{tab:supple_general_image}
    \centering
    \scriptsize
    \resizebox{\linewidth}{!}{%
    \renewcommand{\arraystretch}{1.15}
    \begin{tabular}{c l|ccccc}
        \toprule
        \textbf{Category} & \multicolumn{1}{c|}{\textbf{Methods}} & \textbf{FID}$\downarrow$ & \textbf{pFID}$_{256}\downarrow$ & \textbf{pFID}$_{512}\downarrow$ & \textbf{CLIP-S}$\uparrow$ & \textbf{CLIP-A}$\uparrow$ \\
        \midrule
        \multirow{5}{*}{Natural} & SD Inpainting~\cite{stable_diffusion} & 36.80 & 79.56 & 49.67 & 0.2317 & 5.4230 \\
        & PowerPaint~\cite{powerpaint} & 34.48 & 78.74 & 48.09 & 0.2338 & 5.3168 \\
        & ProOut~\cite{proout} & 29.61 & 78.29 & 46.49 & \textbf{0.2362} & 5.6158 \\
        & Ours (w/o layout) & \textbf{27.43} & \underline{75.46} & \textbf{45.20} & \underline{0.2357} & \textbf{5.6402} \\
        & \textbf{Ours (w/ layout)} & \underline{28.02} & \textbf{75.05} & \underline{46.42} & 0.2354 & \underline{5.6356} \\
        \midrule
        \multirow{5}{*}{Indoor} & SD Inpainting~\cite{stable_diffusion} & 42.03 & 91.38 & 54.20 & 0.2394 & 5.1569 \\
        & PowerPaint~\cite{powerpaint} & 36.68 & \textbf{88.79} & 53.63 & 0.2426 & 5.1291 \\
        & ProOut~\cite{proout} & 34.21 & \underline{89.02} & \textbf{52.27} & \textbf{0.2450} & \textbf{5.3388} \\
        & Ours (w/o layout) & \underline{30.77} & 89.55 & \underline{52.97} & 0.2439 & 5.3254 \\
        & \textbf{Ours (w/ layout)} & \textbf{29.33} & 90.95 & 53.20 & \underline{0.2442} & \underline{5.3377} \\
        \midrule
        \multirow{5}{*}{Human} & SD Inpainting~\cite{stable_diffusion} & 60.49 & 116.31 & 77.62 & 0.2275 & 5.4292 \\
        & PowerPaint~\cite{powerpaint} & 52.08 & 113.50 & 71.73 & 0.2295 & 5.3768 \\
        & ProOut~\cite{proout} & 50.01 & 115.18 & 72.69 & \textbf{0.2336} & 5.5654 \\
        & Ours (w/o layout) & \underline{42.98} & \underline{112.20} & \underline{70.23} & 0.2309 & \underline{5.5822} \\
        & \textbf{Ours (w/ layout)} & \textbf{40.82} & \textbf{109.20} & \textbf{68.56} & \underline{0.2313} & \textbf{5.6052} \\
        \bottomrule
    \end{tabular}}
\end{table}

Table~\ref{tab:supple_general_image} evaluates generalization beyond artwork images without retraining.
We use 400 Open Images samples for each category, namely natural, indoor, and human-centric scenes.
Across all three categories, our method improves FID over ProOut and remains competitive in CLIP-based metrics.

\begin{table}
    \caption{\textbf{DiT-Based Outpainting Evaluation.}
    We evaluate whether our framework transfers to a DiT-based backbone using an SD3-Medium~\cite{sd3} ControlNet inpainting model in the large-canvas setting.
    The final SD3-based model is trained for 20K steps and substantially improves over both progressive SD3 generation and blueprint-based forward initialization.
    }
    \label{tab:supple_dit}
    \centering
    \scriptsize
    \resizebox{0.85\linewidth}{!}{%
    \renewcommand{\arraystretch}{1.15}
    \begin{tabular}{l|ccccc}
        \toprule
        \multicolumn{1}{c|}{\textbf{Methods}} & \textbf{FID}$\downarrow$ &
        \textbf{pFID}$_{256}\downarrow$ & \textbf{pFID}$_{512}\downarrow$ & \textbf{CLIP-S}$\uparrow$ & \textbf{CLIP-A}$\uparrow$ \\
        \midrule
        SD3 Progressive & 138.21 & 135.62 & 132.32 & 0.1827 & 5.5366 \\
        SD3 + Forward Init & 101.59 & 125.25 & 120.87 & 0.1950 & 5.4928 \\
        \textbf{Ours (SD3, 20K steps)} & \textbf{88.56} & \textbf{111.13} & \textbf{113.61} & \textbf{0.1959} & \textbf{6.2075} \\
        \bottomrule
    \end{tabular}}
\end{table}

Table~\ref{tab:supple_dit} evaluates whether the proposed framework transfers to a DiT-based backbone.
We use an SD3-Medium ControlNet inpainting model in the native-1024 setting and evaluate on the large-canvas data used in Table~\ref{tab:supple_arbitrary_canvas}.
Progressive SD3 generation applies the model sequentially, following the progressive outpainting setting.
SD3 with forward initialization uses the Stage~1 blueprint to initialize parallel Stage~2 synthesis without additional training, already improving FID substantially over progressive SD3 generation.
After training our framework with the SD3 backbone for 20K steps, the final model further improves FID, pFID, CLIP-S, and CLIP-A.
These results suggest that our framework can transfer to DiT-based backbones.

\begin{table}
    \caption{\textbf{Inference Speed Comparison.}
        Although our full two-stage model is slower than the simpler progressive baselines on a single GPU, it is already faster than ProOut, which is also designed for high-resolution outpainting.
        Moreover, because our Stage~2 synthesis is parallelizable across patches, the inference time decreases substantially with multiple GPUs, yielding the fastest overall performance.
    }
    \label{tab:supple_speed}
    \centering
    \scriptsize
    \resizebox{0.95\linewidth}{!}{%
    \renewcommand{\arraystretch}{1.2}
    \begin{tabular}{l|ccccccc}
        \toprule
        \textbf{Metric}
        & \textbf{SD Inpainting}
        & \textbf{PowerPaint}
        & \textbf{ProOut}
        & \textbf{Ours}$_{1}$
        & \textbf{Ours}$_{2}$
        & \textbf{Ours}$_{4}$
        & \textbf{Ours}$_{8}$ \\
        \midrule
        Inference Time $\mu$ (s)$\downarrow$
        & 13.4330 & 13.9856 & 18.3108 & 15.7337 & 10.2599 & 7.7082 & 7.5038 \\
        \bottomrule
    \end{tabular}}
\end{table}
Table~\ref{tab:supple_speed} compares the inference time of our method and the baselines.
On a single GPU, our full two-stage model is slower than the simpler progressive baselines, reflecting the additional cost of explicit global planning and guidance.
This single-GPU setting also limits the practical benefit of our parallel Stage~2 design, since all local patches must still be processed on the same device.
However, it is already faster than ProOut, which is designed for high-resolution artwork outpainting.
More importantly, unlike sequential progressive methods, our Stage~2 synthesis can be executed in parallel across patches.
As the number of GPUs increases, the inference time decreases substantially, showing that the main efficiency advantage of our framework comes from its parallel high-resolution generation design.

\begin{table}
    \caption{\textbf{Unfiltered-Caption Robustness.}
        We reevaluate the models using the original BLIP captions without removing generic style prefixes such as ``a painting of''.
        Our method remains stronger than ProOut both with and without layout conditions, and the two settings produce similar image-quality metrics.
    }
    \label{tab:supple_unfiltered}
    \centering
    \scriptsize
    \resizebox{0.8\linewidth}{!}{%
    \renewcommand{\arraystretch}{1.2}
    \begin{tabular}{l|ccccc}
        \toprule
        \textbf{Methods} & \textbf{FID}$\downarrow$ & \textbf{pFID}$_{256}$$\downarrow$ & \textbf{pFID}$_{512}$$\downarrow$ & \textbf{CLIP-S}$\uparrow$ & \textbf{CLIP-A}$\uparrow$ \\
        \midrule
        ProOut & 11.6236 & 9.5937 & 8.0034 & 0.2318 & 6.7972 \\
        Ours (w/o layout condition) & 9.7608 & 9.0624 & 6.7563 & 0.2295 & 6.7886 \\
        Ours (w/ layout condition) & 9.3721 & 9.0603 & 6.7347 & 0.2299 & 6.7956 \\
        \bottomrule
    \end{tabular}}
\end{table}
Table~\ref{tab:supple_unfiltered} reevaluates the models using the original BLIP captions without removing generic style prefixes.
We include this comparison because our default evaluation removes phrases such as ``a painting of'' from the captions of test dataset, whereas ProOut keeps the BLIP test captions unchanged and applies this filtering only to the training data.
To check whether this difference could affect fairness, we repeat the evaluation under the unfiltered test-caption setting.
Under this setting, our method remains stronger than ProOut both with and without layout conditions, showing that the main ranking is robust to the caption preprocessing.

\begin{table}
    \caption{\textbf{Guidance-Scale Sensitivity.}
        Increasing the guidance scale beyond the default value of 3.0 consistently degrades image quality, while the changes in layout accuracy remain small and inconsistent.
        These results suggest that our default setting provides the best overall trade-off.
    }
    \label{tab:supple_guidance_scale}
    \centering
    \scriptsize
    \resizebox{0.8\linewidth}{!}{%
    \renewcommand{\arraystretch}{1.2}
    \begin{tabular}{l|ccccc|cc}
        \toprule
        \multicolumn{1}{c|}{\multirow{2}{*}{\textbf{Methods}}} & \multicolumn{5}{c|}{\textbf{Image
        Quality}} & \multicolumn{2}{c}{\textbf{Layout Acc.}} \\
        \cmidrule(lr){2-6} \cmidrule(lr){7-8}
        & \textbf{FID}$\downarrow$ & \textbf{pFID}$_{256}$$\downarrow$ & \textbf{pFID}$_{512}$$
    \downarrow$ & \textbf{CLIP-S}$\uparrow$ & \textbf{CLIP-A}$\uparrow$ & \textbf{AP}$\uparrow$ &
    \textbf{IoU}$\uparrow$ \\
        \midrule
        Ours$_{7.5}$  & 12.9130 & 10.2372 & 8.3499 & 0.2034 & 6.4015 & 0.4292 & 0.6417 \\
        Ours$_{5.0}$ & 10.4555 & 9.4241 & 7.2545 & 0.2035 & 6.6553 & 0.4274 & 0.6432 \\
        \midrule
        Ours$_{3.0}$ & 9.3064 & 9.0635 & 6.7514 & 0.2033 & 6.7949 & 0.4336 & 0.6382 \\
        \bottomrule
    \end{tabular}}
\end{table}
Table~\ref{tab:supple_guidance_scale} evaluates the effect of the classifier-free guidance scale in our final outpainting model.
As the guidance scale increases from 3.0 to 5.0 and 7.5, the image-quality metrics are consistently degraded.
In contrast, the changes in AP and IoU are relatively small and do not show a consistent improvement trend.
These results indicate that increasing the guidance scale beyond the default setting does not provide a better outputs, and that 3.0 offers the best balance between image quality and layout faithfulness.

\begin{table}
    \caption{\textbf{Forward-Init Strength Sensitivity.}
        Changing the forward-init strength induces a trade-off across metrics.
        While weaker initialization improves full-image and layout-related performance, the default value of 1.0 gives better patch-level fidelity and aesthetic quality, suggesting that alternative settings may also be useful depending on the target objective.
    }
    \label{tab:supple_forward_init}
    \centering
    \scriptsize
    \resizebox{0.8\linewidth}{!}{%
    \renewcommand{\arraystretch}{1.2}
    \begin{tabular}{c|ccccc|cc}
        \toprule
        \multicolumn{1}{c|}{\multirow{2}{*}{\textbf{Methods}}} & \multicolumn{5}{c|}{\textbf{Image
        Quality}} & \multicolumn{2}{c}{\textbf{Layout Acc.}} \\
        \cmidrule(lr){2-6} \cmidrule(lr){7-8}
        & \textbf{FID}$\downarrow$ & \textbf{pFID}$_{256}$$\downarrow$ & \textbf{pFID}$_{512}$$
    \downarrow$ & \textbf{CLIP-S}$\uparrow$ & \textbf{CLIP-A}$\uparrow$ & \textbf{AP}$\uparrow$ &
    \textbf{IoU}$\uparrow$ \\
        \midrule
        0.50 & 9.0319 & 9.2265 & 6.9000 & 0.2051 & 6.7739 & 0.4979 & 0.7013 \\
        0.75 & 9.2196 & 9.1381 & 6.8197 & 0.2037 & 6.7822 & 0.4576 & 0.6619 \\
        \midrule
        1.0 & 9.3064 & 9.0635 & 6.7514 & 0.2033 & 6.7949 & 0.4336 & 0.6382 \\
        \bottomrule
    \end{tabular}}
\end{table}
Table~\ref{tab:supple_forward_init} shows that changing the forward-init strength induces a clear trade-off across metrics.
Weaker initialization improves several full-image and layout-related metrics, whereas the default setting of 1.0 yields better patch-level fidelity and aesthetic quality.
In our main experiments, we use the commonly adopted default value of 1.0, but these results suggest that alternative settings may also be beneficial depending on whether one prioritizes global structural alignment or fine local detail refinement.

\begin{table}
    \caption{\textbf{Parallel vs. Progressive Generation.}
        Replacing the parallel Stage~2 synthesis with a progressive version degrades both image quality and layout accuracy.
        This shows that our parallel generation strategy is not merely an efficiency improvement, but also helps preserve global coherence and final outpainting quality.
    }
    \label{tab:supple_parallel_progressive}
    \centering
    \scriptsize
    \resizebox{0.8\linewidth}{!}{%
    \renewcommand{\arraystretch}{1.2}
    \begin{tabular}{l|ccccc|cc}
        \toprule
        \multicolumn{1}{c|}{\multirow{2}{*}{\textbf{Methods}}} & \multicolumn{5}{c|}{\textbf{Image
        Quality}} & \multicolumn{2}{c}{\textbf{Layout Acc.}} \\
        \cmidrule(lr){2-6} \cmidrule(lr){7-8}
        & \textbf{FID}$\downarrow$ & \textbf{pFID}$_{256}$$\downarrow$ & \textbf{pFID}$_{512}$$
    \downarrow$ & \textbf{CLIP-S}$\uparrow$ & \textbf{CLIP-A}$\uparrow$ & \textbf{AP}$\uparrow$ &
    \textbf{IoU}$\uparrow$ \\
        \midrule
        Ours (progressive)  & 10.0136 & 9.1881 & 7.0942 & 0.2031 & 6.6527 & 0.4137 & 0.6268 \\
        Ours (parallel) & 9.3064 & 9.0635 & 6.7514 & 0.2033 & 6.7949 & 0.4336 & 0.6382 \\
        \bottomrule
    \end{tabular}}
\end{table}
Table~\ref{tab:supple_parallel_progressive} compares our default parallel Stage~2 synthesis with a progressive variant using the same model.
Specifically, the progressive variant follows a ProOut-style sequential procedure, repeatedly regenerating the global blueprint and then synthesizing the corresponding local patch at each step, rather than generating all local patches in parallel from a single shared blueprint.
The progressive variant yields worse image-quality metrics and lower layout accuracy than the parallel generation.
These results show that the benefit of our parallel design is not limited to faster inference, but also includes better preservation of global consistency during high-resolution artwork outpainting.

\section{Extended Baseline Comparison}
\label{sec:supple_baseline}

\begin{table}
    \caption{\textbf{Extended Baseline Comparison.}
        We further examine baseline checkpoint selection by testing the official pretrained PowerPaint model and longer-trained ProOut checkpoints.
        Although the pretrained PowerPaint model was optimized with richer conditioning signals and multiple generation tasks, it does not transfer well to artwork outpainting, while extending ProOut training still does not close the gap to our method.
    }
    \label{tab:supple_extended_baseline}
    \centering
    \scriptsize
    \resizebox{0.8\linewidth}{!}{%
    \renewcommand{\arraystretch}{1.2}
    \begin{tabular}{l|ccccc}
        \toprule
        \textbf{Methods} & \textbf{FID}$\downarrow$ & \textbf{pFID}$_{256}$$\downarrow$ & \textbf{pFID}$_{512}$$\downarrow$ & \textbf{CLIP-S}$\uparrow$ & \textbf{CLIP-A}$\uparrow$ \\
        \midrule
        PowerPaint (pretrained) & 96.6027 & 32.7000 & 52.9863 & 0.1982 & 5.2740 \\
        PowerPaint (40 epoch)  & 10.5374 & 9.4403 & 7.4834 & 0.2003 & 6.4423 \\
        \midrule
        ProOut (30 epoch) & 10.4426 & 9.2552 & 7.3075 & 0.2047 & 6.8305 \\
        ProOut (40 epoch) & 10.2848 & 9.2897 & 7.3197 & 0.2047 & 6.8169 \\
        ProOut (50 epoch) & 10.3032 & 9.2854 & 7.2515 & 0.2052 & 6.8585 \\
        ProOut (60 epoch) & 10.4063 & 9.2894 & 7.3561 & 0.2056 & 6.8573 \\
        \midrule
        Ours & 9.3064 & 9.0635 & 6.7514 & 0.2033 & 6.7949 \\
        \bottomrule
    \end{tabular}}
\end{table}
Table~\ref{tab:supple_extended_baseline} further examines whether the main results depend on baseline checkpoint selection.
We additionally evaluate the official pretrained PowerPaint model because it is trained with richer prompt types, including contextual, object, and shape prompts, and is optimized for multiple tasks beyond outpainting.
As expected, its performance is limited under the substantial domain gap between the pretraining setup and artwork outpainting.
For ProOut, extending training beyond the main setting still does not surpass our method, suggesting that the main results are not explained by baseline checkpoint selection.

\begin{table}
    \caption{\textbf{Baseline AAM Refinements.}
        Applying AAM refinement to progressive baselines is costly and does not improve their overall performance consistently.
        In contrast, our method applies AAM once during global blueprint generation, improving quality while requiring much less additional inference time.
    }
    \label{tab:baseline_aam}
    \centering
    \scriptsize
    \resizebox{0.8\linewidth}{!}{%
    \renewcommand{\arraystretch}{1.2}
    \begin{tabular}{l|ccccc|c}
        \toprule
        \textbf{Methods} & \textbf{FID}$\downarrow$ & \textbf{pFID}$_{256}$$\downarrow$ & \textbf{pFID}$_{512}$$\downarrow$ & \textbf{CLIP-S}$\uparrow$ & \textbf{CLIP-A}$\uparrow$ & \textbf{Speed} (s) \\
        \midrule
        SD Inpainting (w/o AAM) & 10.7598 & 9.5336 & 7.6219 & 0.2007 & 6.7041 & 13.43 \\
        SD Inpainting (w/ AAM) & 10.9945 & 9.2098 & 7.0835 & 0.2022 & 6.7381 & 51.9276 \\
        \midrule
        PowerPaint (w/o AAM) & 10.5374 & 9.4403 & 7.4834 & 0.2003 & 6.4423 & 13.99 \\
        PowerPaint (w/ AAM) & 11.7749 & 9.2007 & 7.1252 & 0.2011 & 6.6010 & 45.7258 \\
        \midrule
        ProOut (w/o AAM) & 10.3032 & 9.2854 & 7.2515 & 0.2052 & 6.8585 & 18.31 \\
        ProOut (w/ AAM) & 11.4558 & 9.1981 & 7.1015 & 0.2051 & 6.8866 & 67.8868 \\
        \midrule
        Ours (w/ AAM) & 9.3064 & 9.0635 & 6.7514 & 0.2033 & 6.7949 & 7.50 \\
        \bottomrule
    \end{tabular}}
\end{table}
Table~\ref{tab:baseline_aam} evaluates an AAM refinement from AlignNoise~\cite{alignnoise} on representative baselines.
Inference time is averaged over 100 random test samples.
We include this comparison because the main results apply AAM only to our method, and we therefore test whether the same refinement can also benefit the baselines under a matched setting.
Our method applies AAM once during global blueprint generation, so the optimization is performed on a single planning path before the final high-resolution synthesis.
In contrast, progressive baselines do not have such a planning stage, and therefore require the refinement to be repeated for each patch generation.
This leads to a substantial increase in inference time, yet does not improve the baseline performance consistently and can even degrade overall quality.
These results suggest that AAM is not an effective drop-in refinement for progressive baselines, whereas our blueprint-guided two-stage design makes it both more effective and more efficient.

\section{Qualitative Results}
\label{sec:supple_qualitative}

\begin{figure}
    \centering
    \includegraphics[width=0.8\linewidth]{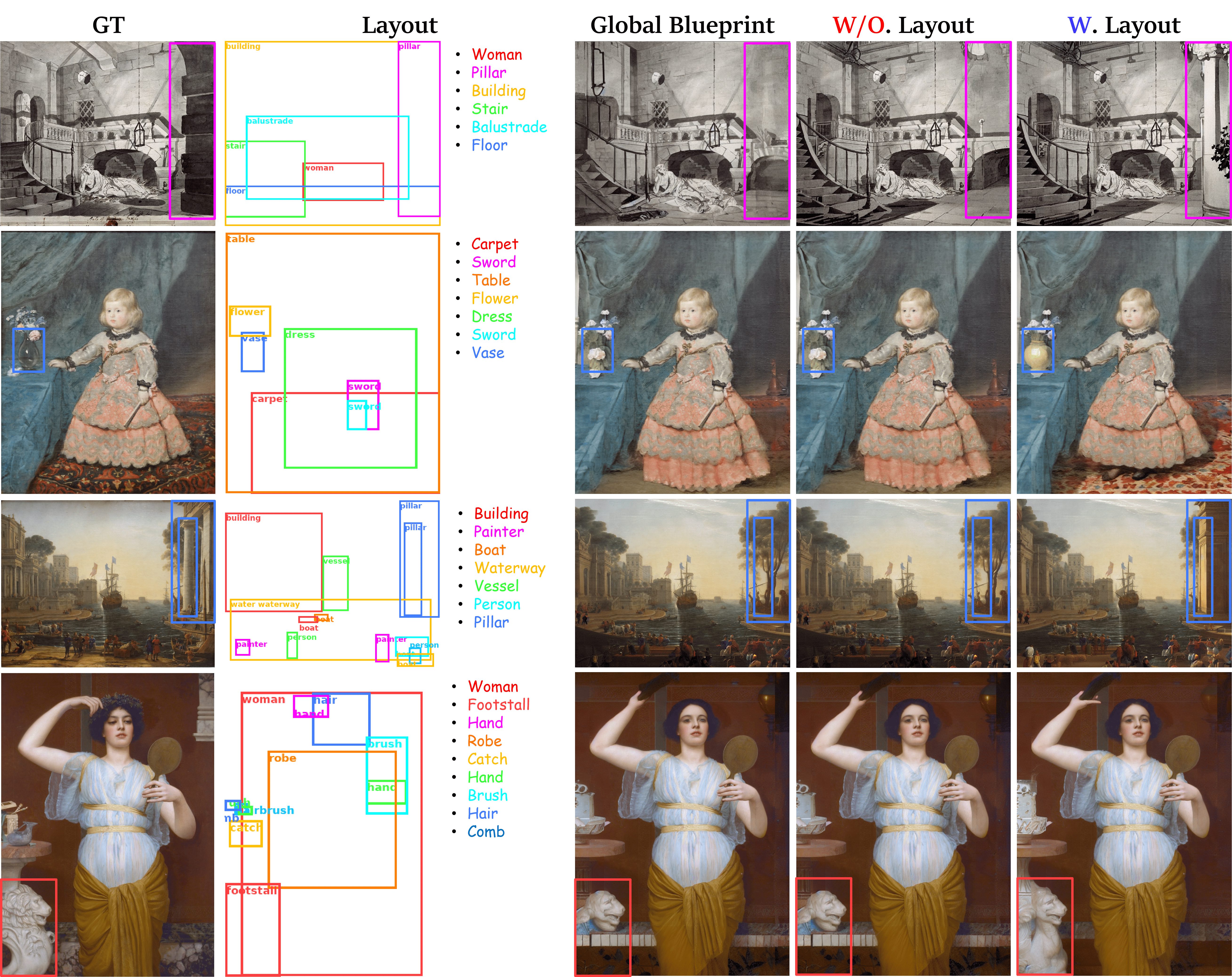}
    \caption{\textbf{Qualitative Ablation on Layout-Conditioned Generation.}
    Our framework produces plausible outpainting results both with and without layout conditions.
    When explicit layout conditions are provided, it more accurately places specified objects at the intended locations, demonstrating improved layout adherence and spatial controllability over the layout-free variant.
    }
    \label{fig:layout_gen}
\end{figure}

\noindent \textbf{Layout-Conditioned Generation.}
We provide additional qualitative comparisons to complement the quantitative results.
Fig.~\ref{fig:layout_gen} compares generation results with and without layout conditions.
In the first three rows, the conditioned objects are located entirely outside the source image, and they are generated only when the layout condition is provided.
The fourth row shows a partially overlapping case, where the object crosses the source boundary, and the outpainted region follows the given condition only in the layout-conditioned result.
Across all examples, both variants produce visually plausible outpainting results, while the layout-conditioned model additionally provides explicit spatial control over object generation in the target region.

\begin{figure}
    \centering
    \includegraphics[width=1\linewidth]{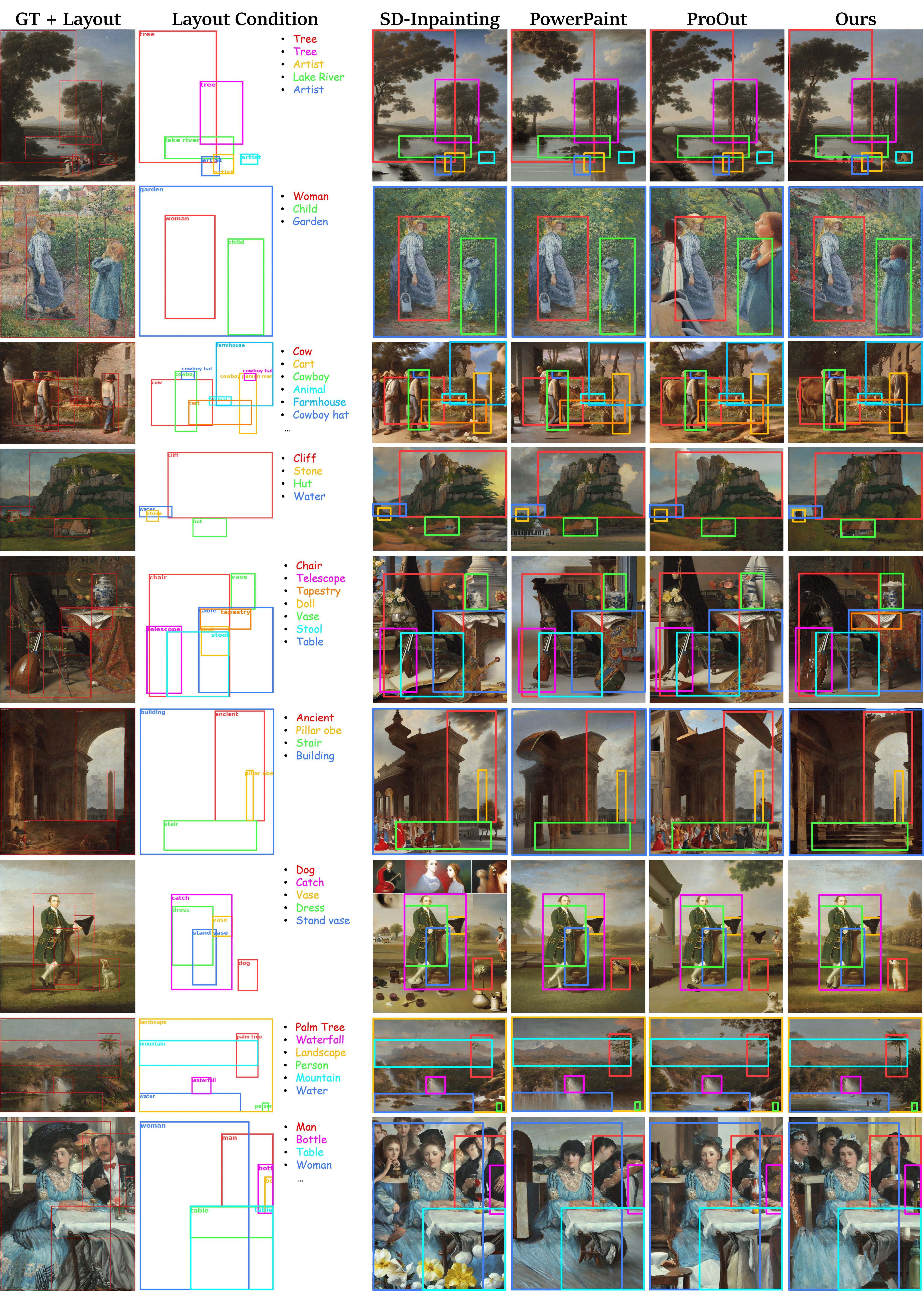}
    \caption{\textbf{Qualitative Comparison of Layout-Guided Generation.}
    Unlike text-only baselines, our framework leverages explicit layout conditions to place specified objects at the intended locations.
    These examples show that layout guidance substantially improves spatial controllability and adherence to the provided object arrangement.
    }
    \label{fig:layout_1}
\end{figure}

Fig.~\ref{fig:layout_1} further compares our method against baselines under the same layout-guided setting.
These additional examples consistently support the observation in the main paper that text-only conditioning is insufficient for precise spatial control, whereas our layout-conditioned framework reliably generates specified objects at the designated locations.

\begin{figure}
    \centering
    \includegraphics[width=0.8\linewidth]{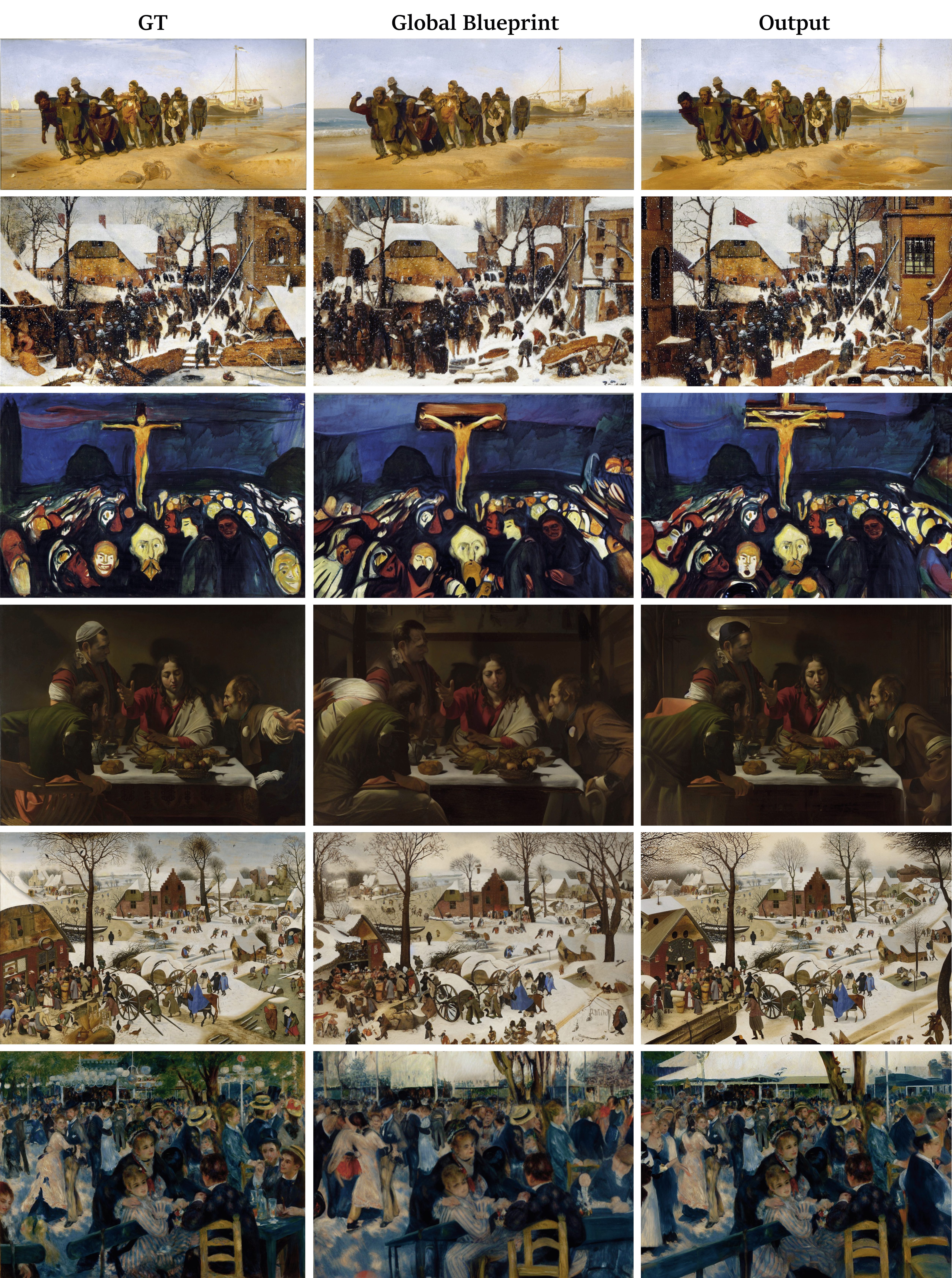}
    \caption{\textbf{Blueprint and Final Output.}
        Visual comparison of the ground-truth images, generated global blueprints, and corresponding final high-resolution outpainting results.
        The final outputs preserve the global semantics and composition established by the blueprints while adding richer local details at high resolution.
    }
    \label{fig:blueprint1}
\end{figure}

\begin{figure}
    \centering
    \includegraphics[width=0.8\linewidth]{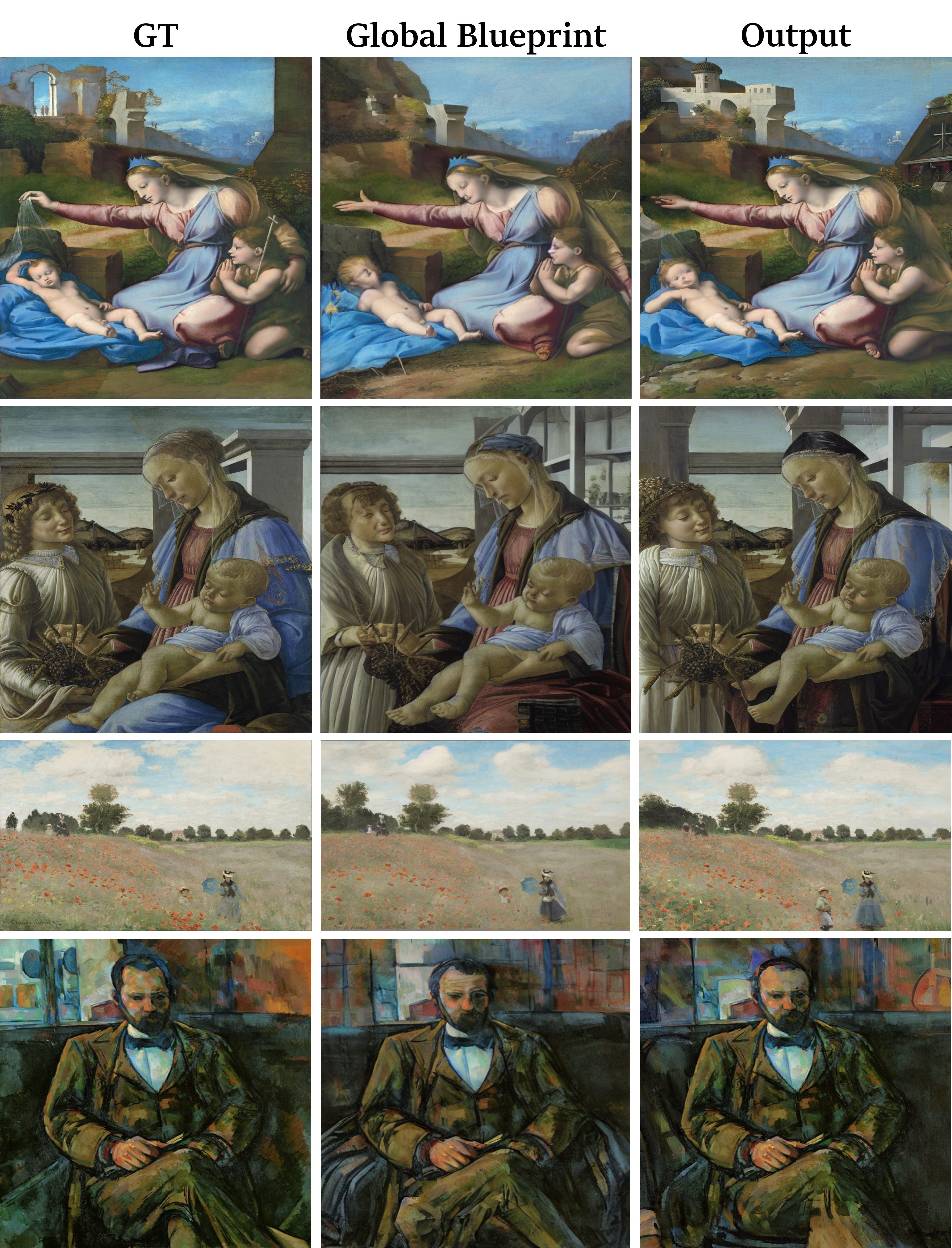}
    \caption{\textbf{Blueprint and Final Output (cont.).}}
    \label{fig:blueprint2}
\end{figure}

\begin{figure}
    \centering
    \includegraphics[width=0.8\linewidth]{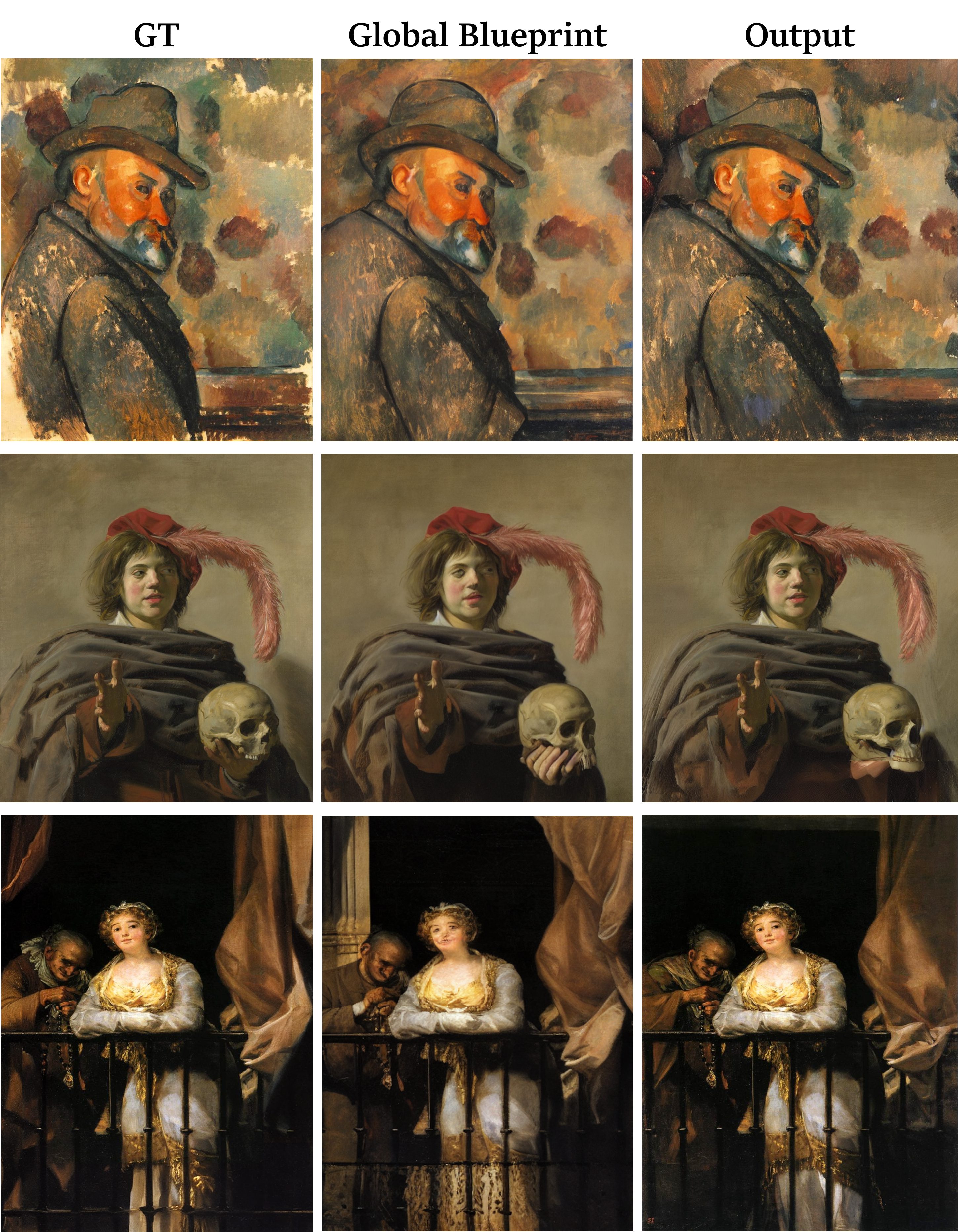}
    \caption{\textbf{Blueprint and Final Output (cont.).}}
    \label{fig:blueprint3}
\end{figure}

\begin{figure}
    \centering
    \includegraphics[width=0.8\linewidth]{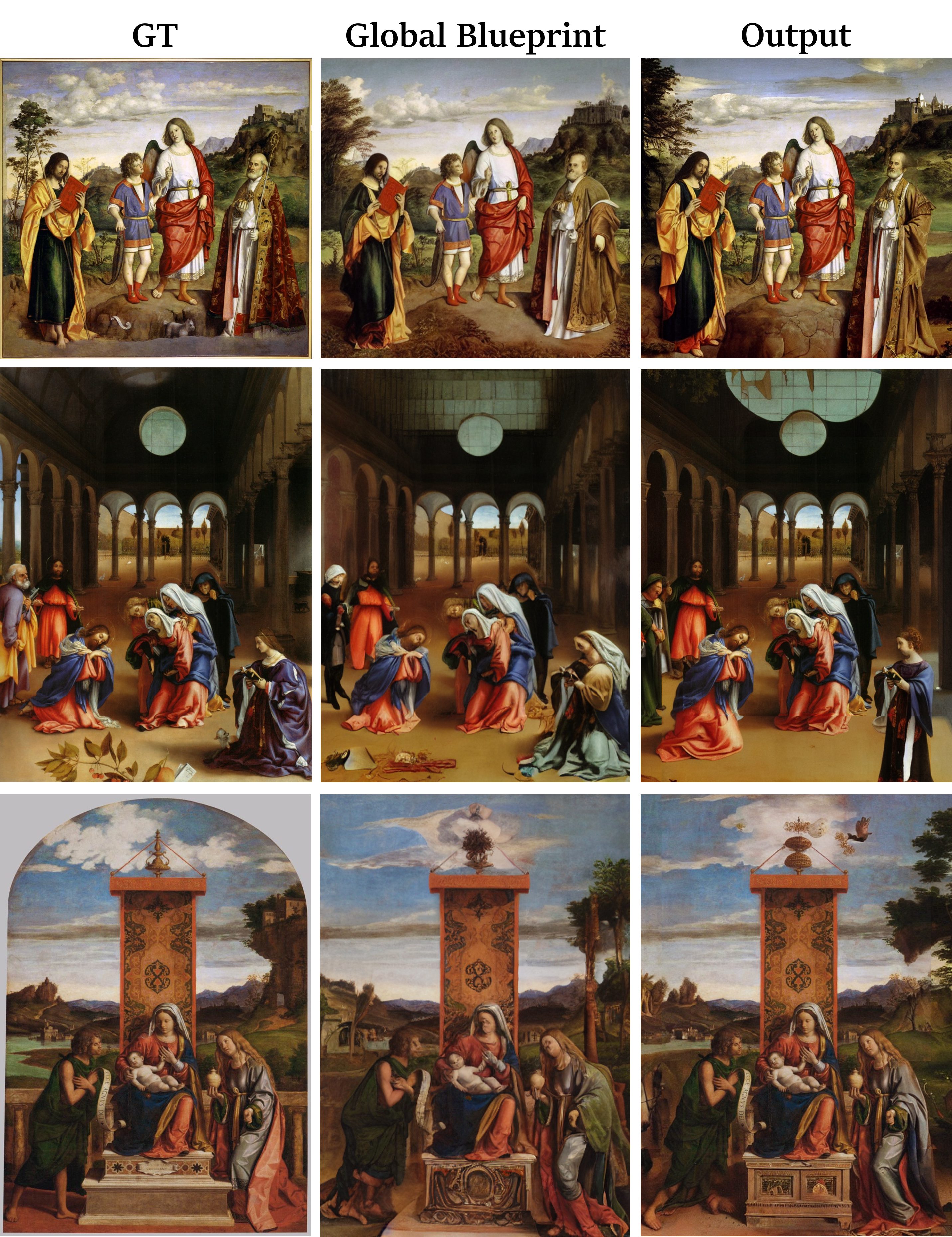}
    \caption{\textbf{Blueprint and Final Output (cont.).}}
    \label{fig:blueprint4}
\end{figure}

\begin{figure}
    \centering
    \includegraphics[width=0.8\linewidth]{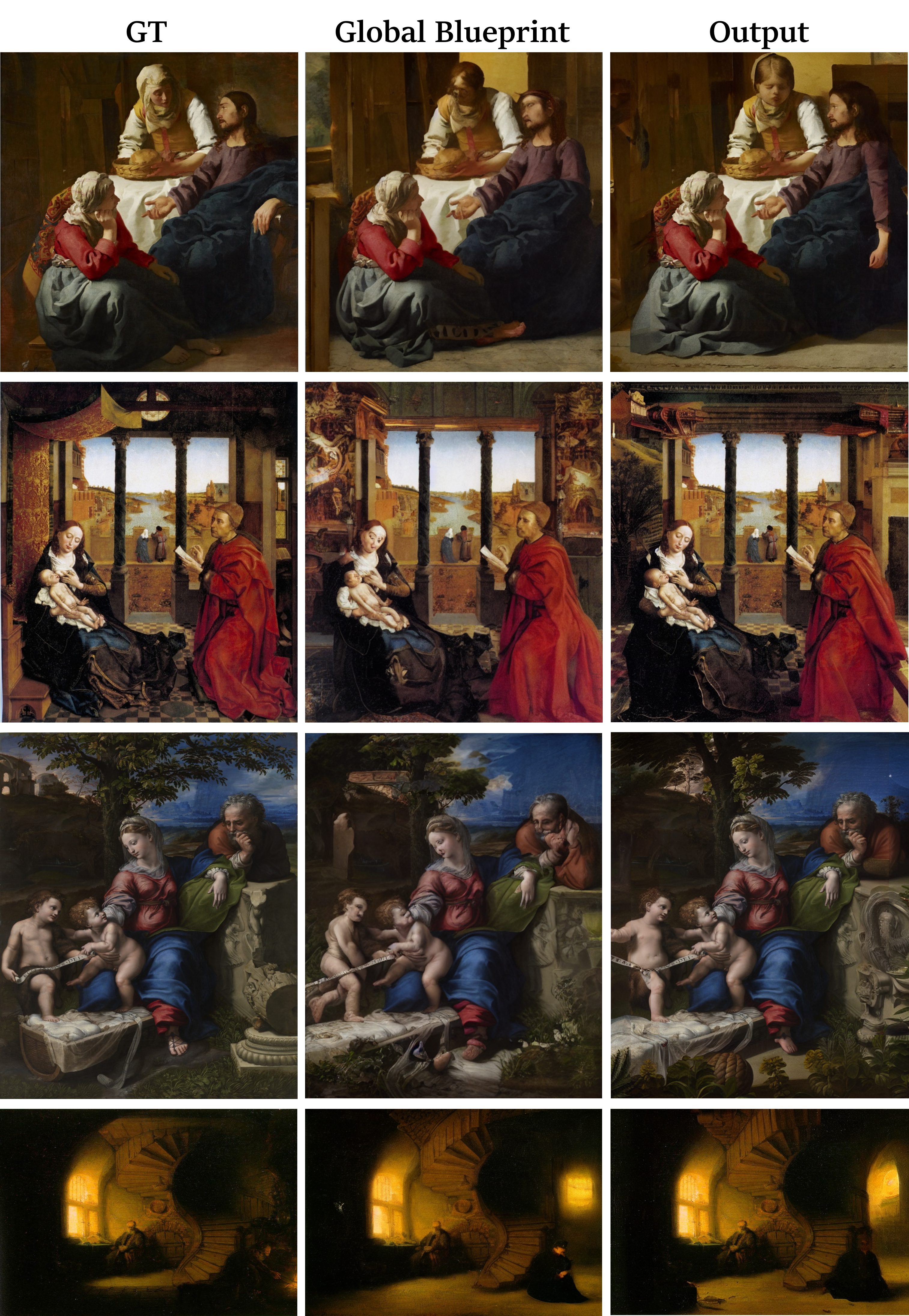}
    \caption{\textbf{Blueprint and Final Output (cont.).}}
    \label{fig:blueprint5}
\end{figure}

\smallskip \noindent \textbf{Blueprint and Final Output.}
Figures~\ref{fig:blueprint1}, \ref{fig:blueprint2}, \ref{fig:blueprint3}, \ref{fig:blueprint4}, and \ref{fig:blueprint5} compare the ground-truth image, the generated global blueprint, and the final outpainting result.
Across diverse examples, the final result consistently follows the overall semantics and object arrangement established in the blueprint, while introducing high-resolution local refinement with richer detail and stylistic consistency.
These examples suggest that the blueprint serves as an effective global guide for the final outpainting stage.

\begin{figure}
    \centering
    \includegraphics[width=1\linewidth]{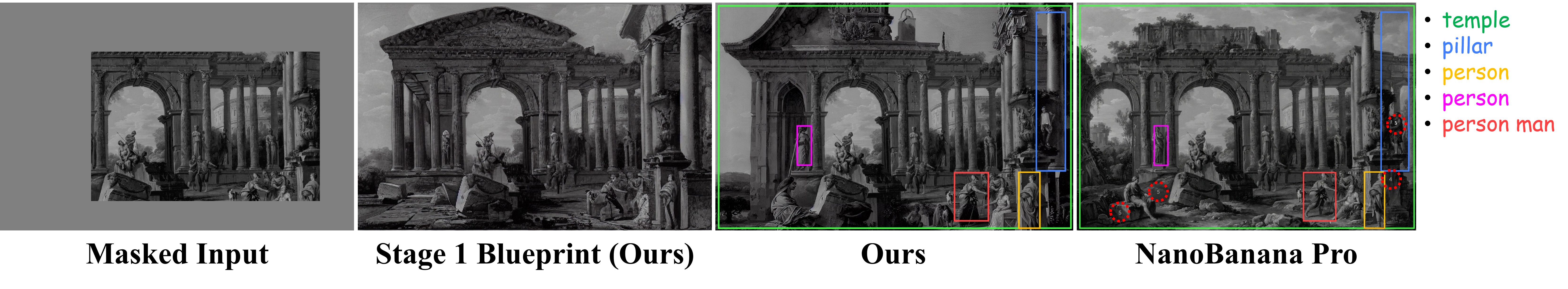}
    \caption{\textbf{Qualitative Comparison with Nano Banana Pro.}
    The figure includes the masked input, the Stage~1 blueprint, our final Stage~2 output, and the Nano Banana Pro~\cite{nanobananapro} output.
    Our method follows the layout more reliably while supporting arbitrary aspect ratios and resolutions through reconfigurable Stage~2 patch tiling.
    }
    \label{fig:nanobanana}
\end{figure}

\smallskip \noindent \textbf{Comparison with Nano Banana Pro.}
Fig.~\ref{fig:nanobanana} compares our method with Nano Banana Pro~\cite{nanobananapro} under a setting with an arbitrary expansion ratio.
The figure shows the masked input, the Stage~1 blueprint, our final Stage~2 result, and the Nano Banana Pro result.
We generated the Nano Banana Pro outputs via Google AI Studio (May 2026).
Since direct layout conditioning for Nano Banana Pro is not straightforward, we provide layout conditions using a layout map image and text prompts containing the global caption and object coordinates.
Nano Banana Pro produces visually plausible results, but it can miss layout conditions or introduce artifacts in the generated regions.
It is also restricted to predefined aspect ratios and resolutions.
In contrast, our method follows the layout more reliably through explicit conditioning in Stage~1 and supports arbitrary aspect ratios and resolutions through reconfigurable Stage~2 patch tiling.

\begin{figure}
    \centering
    \includegraphics[width=1\linewidth]{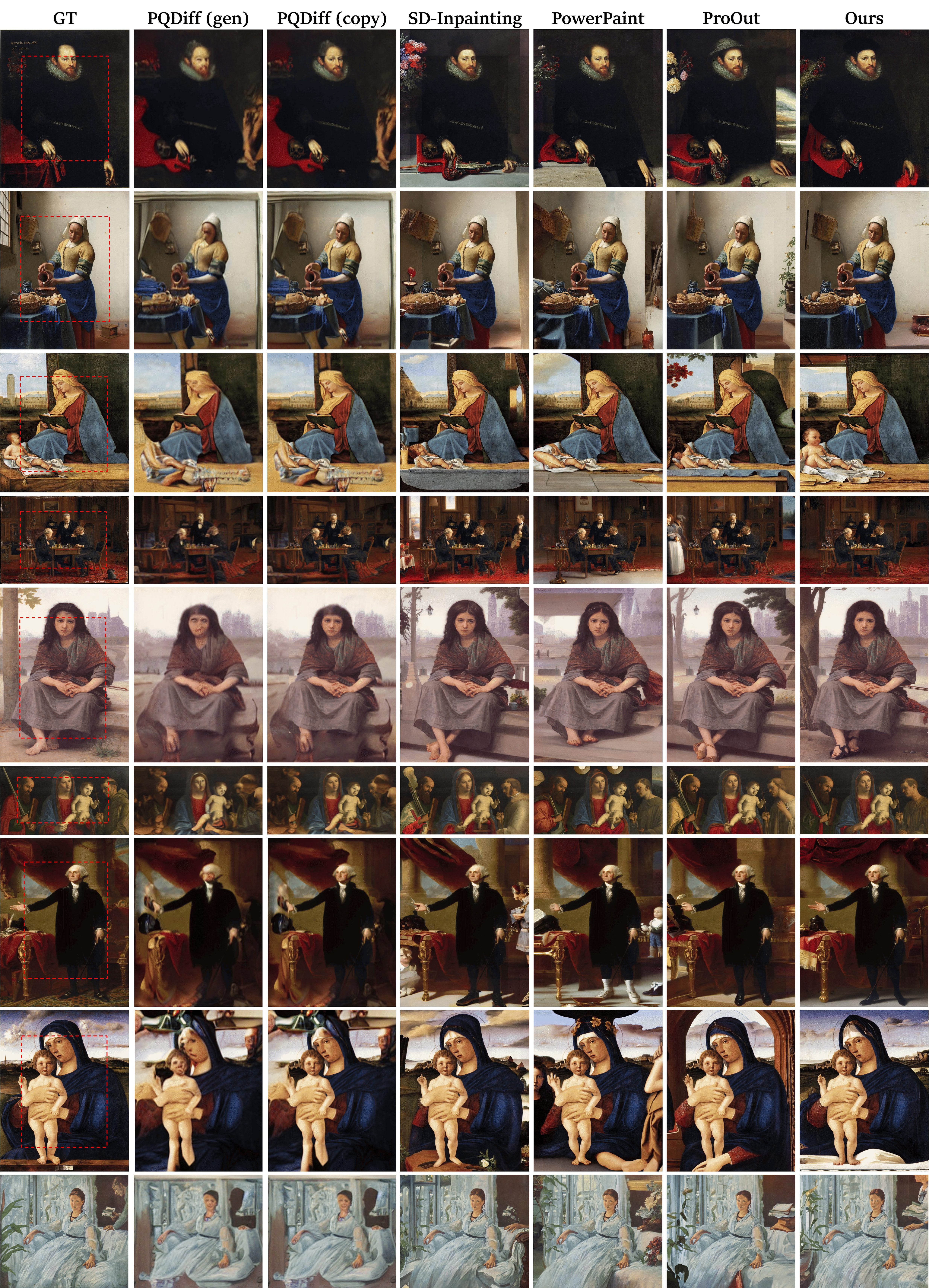}
    \caption{\textbf{Additional Qualitative Results.}
    The red dashed lines in the ground-truth column indicate the boundaries of the original source region.
    Compared with baseline methods, our framework produces more semantically coherent and visually faithful outpainting results, with better preservation of global structure and fine details.
    }
    \label{fig:quali_1}
\end{figure}

\begin{figure}
    \centering
    \includegraphics[width=1\linewidth]{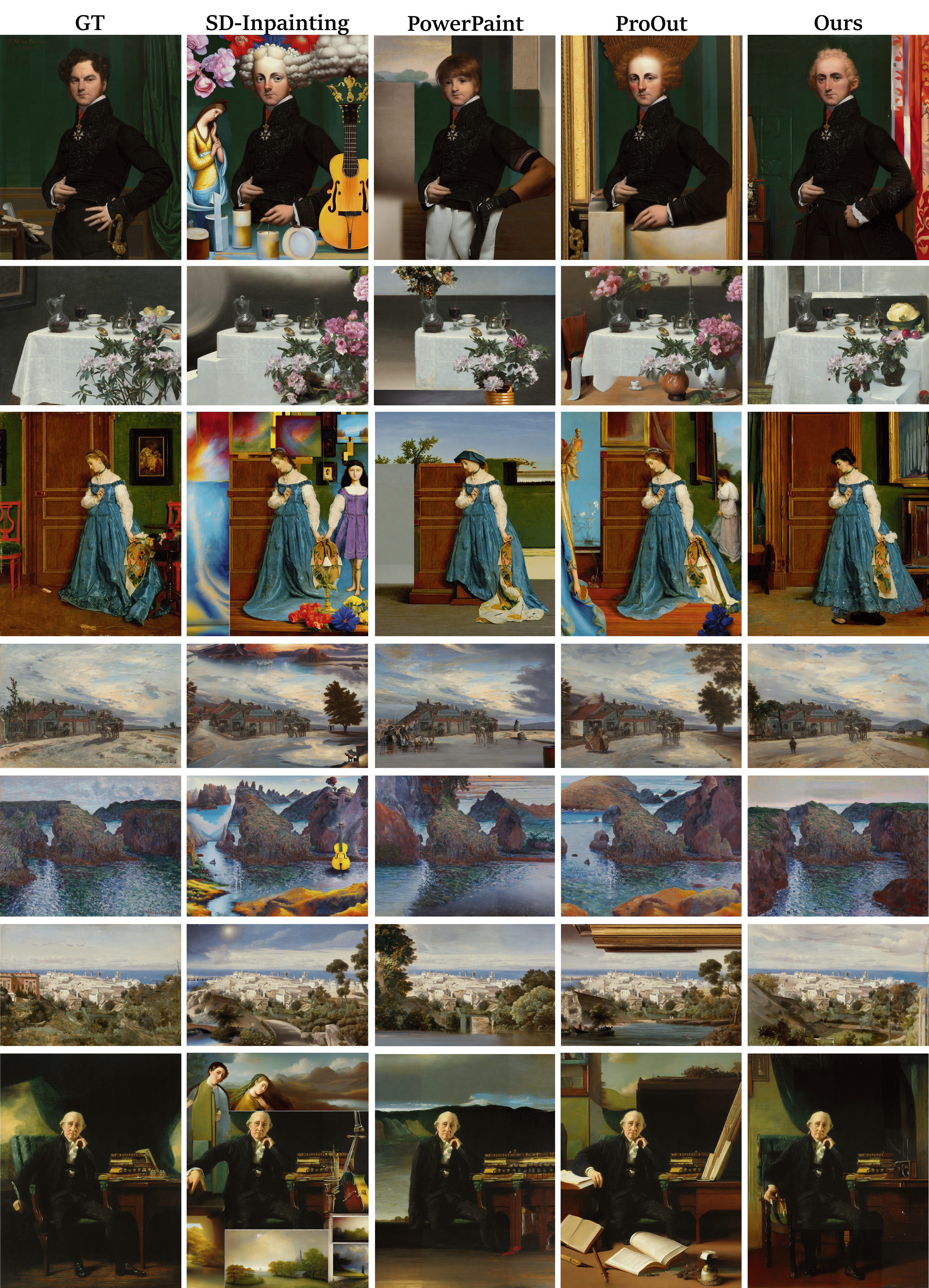}
    \caption{\textbf{Qualitative Results on Landscape Artworks.}
    Additional comparisons on landscape artwork images show that our framework maintains more coherent global composition and stylistic consistency than progressive baselines under wider outpainting settings.
    PQDiff is excluded because it operates at a fixed expansion ratio and therefore cannot accommodate varying canvas sizes.
    }
    \label{fig:quali_scene_data}
\end{figure}

\smallskip \noindent \textbf{Additional Qualitative Comparisons.}
Fig.~\ref{fig:quali_1} and Fig.~\ref{fig:quali_scene_data} present additional qualitative comparisons on the IconArt~\cite{iconart} dataset and landscape artwork images~\cite{cleveland_museum}, respectively.
Across all examples, our method consistently produces semantically coherent and visually faithful results that naturally extend the source content.
In contrast, PQDiff~\cite{pqdiff} suffers from blurriness due to the intermediate super-resolution step, while progressive baselines such as SD Inpainting~\cite{stable_diffusion}, PowerPaint~\cite{powerpaint}, and ProOut~\cite{proout} occasionally exhibit semantic disconnection between the source and generated regions or introduce noticeable artifacts.

\begin{figure}
    \centering
    \includegraphics[width=0.7\linewidth]{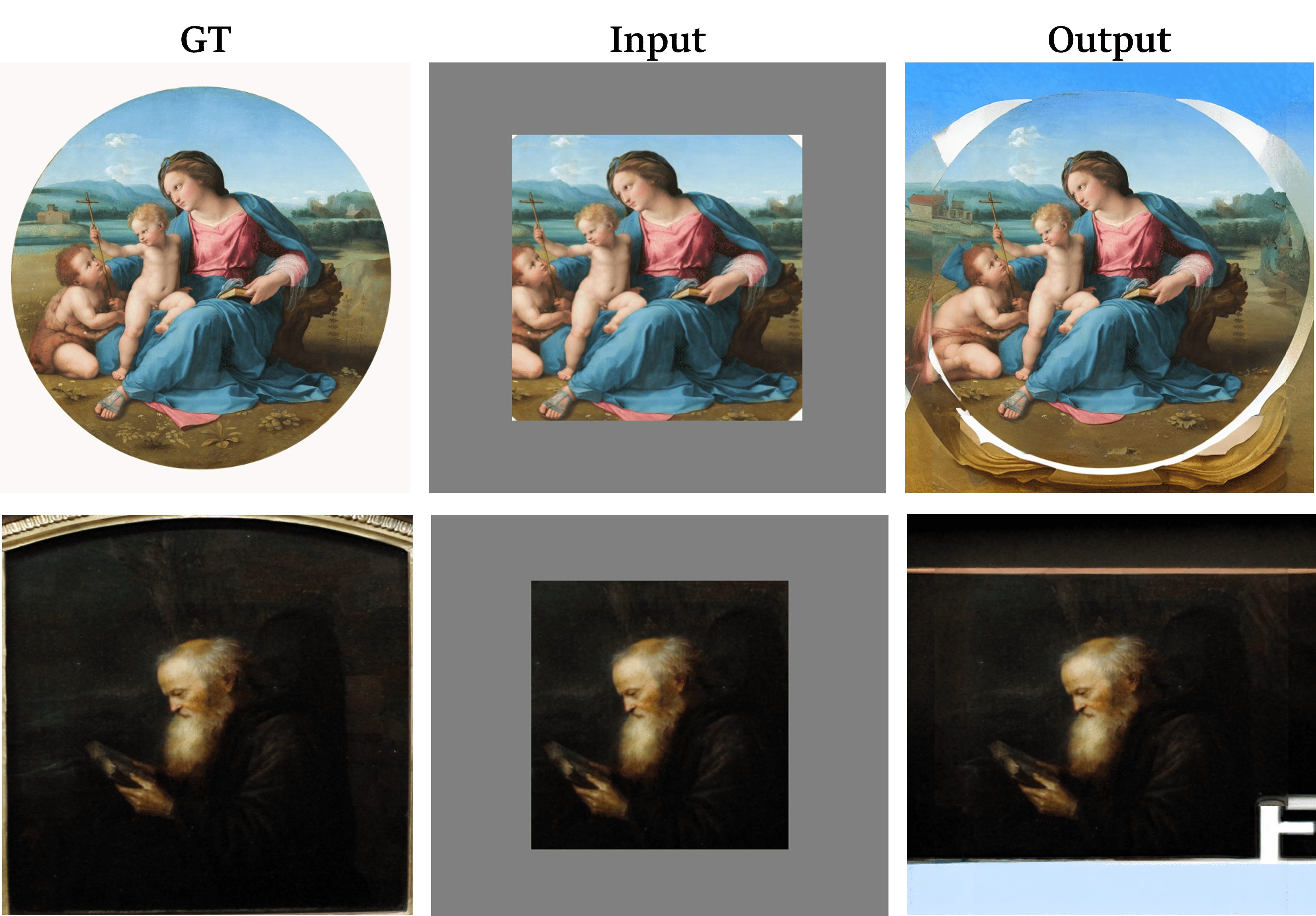}
    \caption{\textbf{Failure Cases.}
    The model may produce unstable outpainting results when the source context near the masking boundary contains visually uniform or monochrome patterns (top).
    Additionally, generation quality may degrade in very dark scenes (bottom), leading to visual artifacts or structural inconsistencies.
    }
    \label{fig:failure_case}
\end{figure}

\smallskip \noindent \textbf{Failure Cases and Limitations.}
Fig.~\ref{fig:failure_case} shows representative limitations of our framework.
When the source context near the masking boundary contains visually uniform or monochrome patterns, the model may receive weak structural cues and produce unstable outpainting results.
We also observe degraded generation quality in very dark scenes, where limited visual information can lead to artifacts or structural inconsistencies.

\end{document}